\documentclass[10pt, a4paper]{article}
\usepackage{comment}
\usepackage[UTF8]{ctex} 
\usepackage{booktabs}

\usepackage[]{lrec2026} 
\usepackage{xcolor}

\title{An Empirical Study on Chinese Character Decomposition in Multiword Expression-Aware Neural Machine Translation}

\name{Lifeng Han$^{1,2}$, Gareth J. F. Jones$^3$,  Alan F. Smeaton$^4$} 

\address{ $^1$Biomedical Data Sciences, Leiden University Medical Center, NL \\ $^2$Leiden Institute of Advanced Computer Science (LIACS), Leiden University, NL\\
$^3$ADAPT Research Centre, Dublin City University, Dublin, Ireland\\
$^4$Insight Centre for Data Analytics, Dublin City University, Dublin, Ireland\\
         l.han@liacs.leidenuniv.nl l.han@lumc.nl\\}

\abstract{
Word meaning, representation, and interpretation play fundamental roles in natural language understanding (NLU), natural language processing (NLP), and natural language generation (NLG) tasks. 
Many of the inherent difficulties in these tasks stem from Multi-word Expressions (MWEs), which complicate the tasks by introducing ambiguity, idiomatic expressions, infrequent usage, and a wide range of variations. Significant effort and substantial progress have been made in addressing the challenging nature of MWEs in Western languages, particularly English. This progress is attributed in part to the well-established research communities and the abundant availability of computational resources.
However, the same level of progress is not true for language families such as Chinese and closely related Asian languages, which continue to lag behind in this regard. 
While sub-word modelling has been successfully applied to many Western languages to address rare words improving phrase comprehension, and enhancing machine translation (MT) through techniques like byte-pair encoding (BPE), it cannot be applied directly to ideograph language scripts like Chinese.
In this work, we conduct a systematic study of the Chinese character decomposition technology in the context of MWE-aware neural machine translation (NMT). 
Furthermore, we report experiments to examine how Chinese character decomposition technology contributes to the representation of the original meanings of Chinese words and characters, and how it can effectively address the challenges of translating MWEs. 
 \\ \newline  }

\begin{document}

\maketitleabstract


\section{Introduction}
\label{intro}
\footnote{the initial draft version of this work was finished by 2022 when LH was at DCU, content extracted from PhD thesis \cite{han2022investigation}}
Word 
meaning representation is a fundamental  research topic in natural language processing (NLP) and natural language understanding (NLU) because of the key focus on the word as a unit and the correct interpretation of 
\textit{word meaning} in natural languages. Thus, the study of word meaning interpretations holds huge importance  for downstream applications such as natural language generation (NLG) including machine translation (MT), and summarisation, among others.

Correct interpretation and representation of combinations of words in multi-word expressions (MWEs) has been a bottleneck to progress in many NLU and NLP tasks \cite{Sag2002MWE}. Indeed most state-of-the-art computational models can not capture and interpret MWEs accurately and reliably because of the idiomatic 
nature of MWEs both in their forms, including unexpected part-of-speech patterns (e.g. by and large, prep+conj+adj), and in their metaphorical use, among others.  

For interpreting word meanings in computational settings, sub-word modelling techniques like byte-pair encoding (BPE) have gained significant popularity in recent years. These techniques have been particularly effective in processing Western alphabetical languages, offering the advantage of fuzzy matching for word meanings. For instance, they can use stem or morpheme components to estimate the meanings of rare words and phrases \cite{SubwordNMT15Sennrich}. However, such technology does not apply directly to ideograph (\textit{aka} ideogram) languages such as Chinese, Japanese, or Korean. 

Fortunately, relevant research work has been carried out in the field of Chinese and Asian language character (\textit{aka} symbol) decomposition and its applications in different NLP tasks. Among these, some promising works have been published on the MT task \cite{HanKuang2018NMT,Park_Zhao2020_Korean_nmt,stratos-2017-sub_korean_nlp,zhang-komachi-2018-neural}.  
Despite this progress, there are a number of important research questions that remain to be addressed, such as how to 
capture  word 
meaning correctly, how to 
cope with the challenges caused by low-frequency or \textit{out-of-vocabulary} (OOV) words and phrases, how to properly integrate \textit{linguistic knowledge}, and how to correctly translate  
idiomatic MWEs \cite{han-etal-2020-alphamwe,han_gladkoff_metaeval_tutorial2022,constant-etal-2017-survey,Savarytv2018vMWE}.

In this work, we report on an empirical study investigating the effects of Chinese character decomposition on  
word meaning representations and MWE acquisition in the context  of MT for
the Chinese $\rightarrow$ English (zh$\rightarrow$en) language pair. We address this from two aspects, i), how and to what degree does Chinese character decomposition represent the original word and character meaning in the context of MT and the translation of MWEs and ii)  whether such decomposition models can be combined with traditional methods such as bilingual MWE (BiMWE) term extraction in order to further boost the performance on MWE translation.


In this  investigation, we first undertake a review of related work carried out in this field including lexical granularity for MT, Chinese character decomposition for NLP, and investigating MWEs in MT. Then we report several experiments investigating the above two research questions including some re-examination of the findings published in the recent work by \cite{HanKuang2018NMT,han-etal-2020-multimwe,han-etal-2020-alphamwe,HanJonesSmeatonBolzoni2021decomposition4mt_MWE}.

The review  covers  research published in both statistical and neural computational models of language processing. However, the empirical investigations and evaluations  focus on neural MT (NMT) models including the bi-directional recurrent neural network (BiRNN) structure and the Transformer structure \cite{google2017attention}. This is due to the fact that, regarding the high-resource and high-performing  Chinese $\rightarrow$ English language pair, NMT has replaced Statistical MT (SMT) as the new state-of-the-art, even though it still struggles to achieve human parity in terms of performance in some domains such as literature text \cite{brown-etal-1993-mathematics,Koehn2010,cho2014encoder-decoder,Google2016MultilingualNMT,han-etal-2020-alphamwe}.

To understand word meaning from a linguistically motivated viewpoint, we carry out an investigation into the composition, form, and philosophy of Chinese characters and their meaning representation at
different decomposition levels. To better understand MWEs in aiding MT performance, we apply an automatic bilingual MWE (BiMWE) extraction framework into the zh$\rightarrow$en language pair and we integrate the extracted bilingual MWE terms back into the training corpus as data augmented with learned 
MWE features. Then we examine the effects of character decomposition methods on BiMWE augmented NMT learning.

\begin{figure*}[!htb]
\begin{center}
\centering
\includegraphics*[width=1\textwidth]{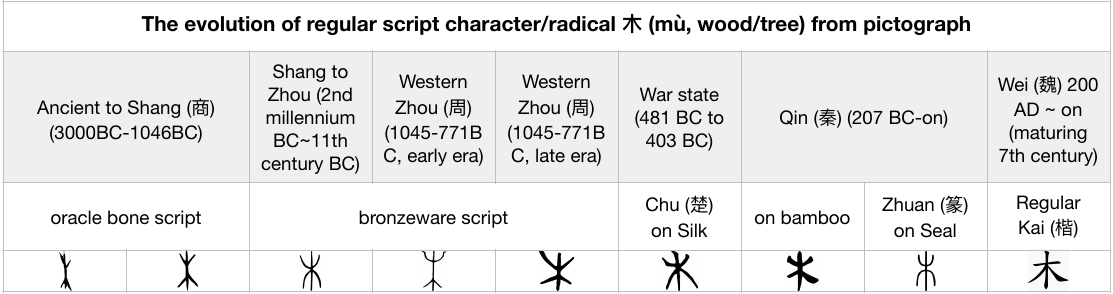}
\caption{Character radical 木 (mù, wood/tree) from pictograph: the root of semantic representation.} 
\label{fig:radical_wood_evolution}
\end{center}
\end{figure*}

\begin{figure*}[]
\begin{center}
\centering
\includegraphics*[width=1\textwidth]{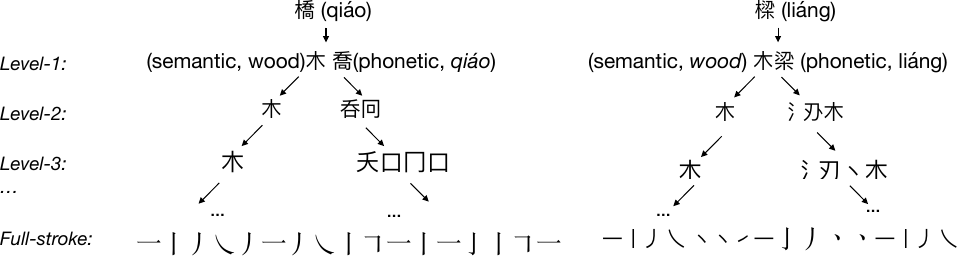}
\caption{Character decomposition examples of qiáo and liáng}
\label{fig:decompose_degree_qiaoliang}
\end{center}
\end{figure*}

For a glimpse of Chinese characters, we look at Figure~\ref{fig:radical_wood_evolution} and Figure~\ref{fig:decompose_degree_qiaoliang} where we present the Chinese character decomposition of ``橋(qiáo)'' and ``樑(liáng)'', and the evolution process of their radical part ``木 (mù)''.

Chinese characters are often composed of two different parts namely a semantic part (意旁\, yì páng) which is also called a radical to indicate partially of the overall character meaning, and a  phonetic part (聲旁\, shēng páng, \textit{aka} 音旁\, yīn páng) which indicates the pronunciation of the overall character.
For the semantic part, there are examples such as what kind of material is this object made of if it is an object (often as a noun), and sometimes a radical is related to an action and indicates the overall character is a verb. For instance, in Figure~\ref{fig:decompose_degree_qiaoliang}, the radical ``木 (mù, wood/tree)'' indicates that historically ``橋(qiáo, bridge)'' and ``樑(liáng, beam)'' were made of wood material. The radical ``木 (mù, wood/tree)'' preserves the meaning of wood and tree because it has evolved from a pictogram (\textit{aka} pictograph) symbol that is a drawing of a tree from 3000BC as shown in Figure~\ref{fig:radical_wood_evolution}.
Some radicals can be independent characters like ``木 (mù, wood/tree)'' and ``金 (jīn, or 釒, metal)'', however, there are also radicals that can not be independent characters, for instance, \, 刂\,(tí dāo páng, knife, sword) and ``艹" (cǎozì tóu, grass) from \cite{HanKuang2018NMT,HanJonesSmeatonBolzoni2021decomposition4mt_MWE}.

For the phonetic part of the Chinese character, there are examples like ``
喬 (qiáo)'' 
and ``梁 (liáng)'' which have the exact same pronunciations as the overall characters they are placed in, i.e. ``橋(qiáo, bridge)'' and ``樑(liáng, beam)''. However, there are also examples like the character 劍\,(jiàn, \emph{sword}) where the phonetic part  ``僉\, (qiān)'' partially indicates the overall character's sound  \cite{HanJonesSmeatonBolzoni2021decomposition4mt_MWE}.
From the example characters in Figure~\ref{fig:decompose_degree_qiaoliang}, we can see that both semantic and phonetic parts of the character can be further decomposed into smaller character pieces or strokes, with a certain ordering.

Prior to this study, little work had been reported which systematically examined the full potential of these alternative levels of decomposition of Chinese characters for use in MT and especially how such decomposition models affect the BiMWE dictionary-based feature learning in NMT models.
Potentially these can  be 
added to the meaning representation in MT,
which addresses some of the translation challenges of MWEs.

In summary, in this empirically-based study, we investigate a number of issues including (i) whether Chinese radicals can enhance word-level NMT learning as semantic features (\textit{pilot-study} $Radical\_Sem$);
(ii) to what degree the Chinese radical and stroke sequences can represent the original meaning of words from which they are decomposed  (\textit{pilot-study} $Radical\_Sem$ and \textit{Model} $Decompose\_Rep$);
 (iii) what differences in translation performance can be observed when using various levels of decomposition for Multiword Expressions (\textit{Model} $Decompose\_Rep$);
 (iv) what are the effects of radical and stroke representations in MWEs for MT (\textit{Model} $Decompose\_Rep$ and \textit{$BiMWE\_Term$}); (v) what are the effects of automatically extracted  BiMWE terms in NMT learning in a rich language resource scenario such as zh$\rightarrow$en on both the character and decomposed levels (\textit{Model} $BiMWE\_Term$). 

This paper is organised as follows: in  Section~\ref{intro} we introduced the topic and scope of the research in more detail; 
Section~\ref{relatedwork_section} presents related work while Section~\ref{section_model_idea} describes the model designs of our systematic investigation; 
Section~\ref{model_eval_section} reports an experimental evaluation of our structured models; 
and in Section~\ref{discussion_conclusions} we present further discussions and conclusions to the work.

\section{Related Work}
\label{relatedwork_section}

We first introduce 
related 
work in MT examining alternative 
input units (\textit{aka} granularity) for statistical and neural sequence learning, mostly focusing on 
Chinese, Japanese, and other ideographic languages.
Following that we then 
introduce work in the area of
Chinese character decomposition 
in other related NLP tasks. Finally, we review 
related work investigating MWEs for MT.


\subsection{Lexical Granularity for MT}
\label{relatedwork_granularity4mt}

\begin{figure*}[htb]
\begin{center}
\centering
\includegraphics*[width=1\textwidth]{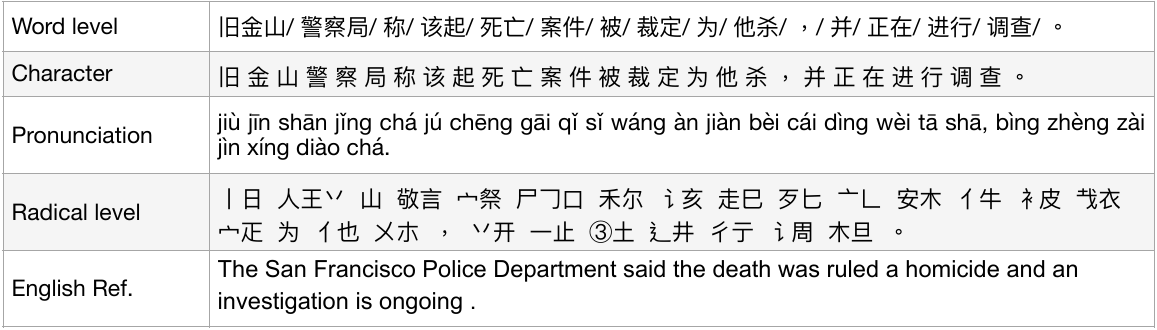}
\caption{Chinese sentence example with word segmentation (delimiter `/'), character sequence, radical-level decomposition and pronunciations.}
\label{fig:word2character_MT}
\end{center}
\end{figure*}

An example of lexical granularity for MT usage using the Chinese language is shown in Figure~\ref{fig:word2character_MT}, where word-level, character-level, and radical-level sequences are placed, together with the character pronunciations and one English reference translation.

To explore the combined performances of
both word- and character-level knowledge, \cite{chen-etal-2018-combining} carried out NMT investigations on Chinese~$\leftrightarrow$~English (zh~$\leftrightarrow$~en) and English~$\rightarrow$~German (en~$\rightarrow$~de) language pairs. The authors argue that the hierarchical combination of word and character models via multi-scale attention performs better than a sub-word encoding BPE baseline on the zh~$\leftrightarrow$~en 
language pair. This 
verifies that the BPE sub-word model does not directly apply to ideographic languages in the way that it can be
applied to Western alphabetical languages.

While character-based NMT models have shown better performances than word-level, 
\textit{OOV} and \textit{unseen words} translation still presents challenges and remains unsolved, as reflected in the work by \cite{dai-yamaguchi-2019-compact} on Japanese-English NMT where the authors explored methods including a mid-gated model and restricting beam search sizes for better performance. This inspires us to further explore one level deeper than characters, by using Chinese character decomposition methods to investigate better semantic modelling for rare word translation.

An input of character sequences is also compared to word-level sequences for MT evaluation tasks and this shows the advantages of having
word boundaries in languages such as Chinese, where there are 
meaningful internal structures inside the words
\cite{liu-ng-2012-character}.

In order to build a dictionary for multilingual MT that can be shared between both close (e.g. Catalan / Spanish) and distant languages (e.g. Turkish / English / Chinese) in terms of their alphabets, \cite{costa-jussa-etal-2017-byte} carried out a comparison between byte-based and character-based NMT. While the testing scores achieved were similar, the byte-based model reduced the training time for model learning. This leads 
us to explore 
performance optimisation relating to different encoding sequences, such as the training cost and the hardware resource requirements, in addition to the translation evaluation scores. 

Other inspirational sub-word or single-character level computational models proposed for Western language pairs for MT include work in \cite{beinborn-etal-2013-cognate} which studied 
cognate production using a character-based SMT model.
Cognates are words in different languages that are associated with each other by language learners and they demonstrated that a character-based MT model can learn the necessary patterns for producing cognates in different languages and alphabets.
 \cite{costa-jussa-fonollosa-2016-character}  demonstrated the effectiveness of a varied window size of character sequence embeddings in comparison to phrase-based SMT and NMT baselines 
 In other work, \cite{Salesky2020OptimizingSG_nmt} explored the optimisation of segmentation granularity for NMT on morphologically-rich languages namely English~$\rightarrow$~Czech/German (en~$\rightarrow$~cs/de) using BPE,
 and others, e.g. \cite{tiedemann-2012-character}.

All this previous work inspires us to explore  alternative 
input units for Chinese and other ideographic languages to determine which are the best for translation performance
at capturing word meaning in computational modeling, especially handling the translation of MWEs.

Some very closely related work to ours is from \cite{zhang-komachi-2018-neural,ZhangMatsumoto18radical} where the authors also explored deeper level segmentation of Chinese characters into radicals or strokes and also tested their corresponding performances between English, Chinese and Japanese NMT. However, in our work, we not only investigate the Chinese radical as a semantic feature of model learning, but also explore  different decomposition levels from shallow to deep, and the relative performances of these models when 
translating MWEs. We also
carry out some experimental comparisons with this
related work.

\subsection{Chinese Character Decomposition in NLP}
\label{relatedwork_zh_decompose4nlp}

Chinese character decomposition has been investigated in previous work for other NLP tasks besides translation. For instance, \cite{Radical15ShiNLP} examined the incorporation of radical embeddings into short-text categorisations, Chinese word segmentation, and web search ranking. 
The authors suggest that radical-level decomposition should be the deepest level of analysis in Chinese NLP since it is difficult to model 
semantic preservation 
via deeper Chinese strokes. In our experiments, however, our findings suggest that even deeper Chinese character decomposition (e.g to level-3 in Section~\ref{model_eval_section}), 
including strokes, can also achieve good results with the model still able to acquire the meaning of the original Chinese word sequences. 

Radical knowledge has also been applied to Chinese word similarity, Chinese named entity recognition, and parsing and semantic role labelling 
\cite{Dong2016CharacterRadical,Cao2018cw2vecLC,Wu2019GlyceGF}. For instance, \cite{Dong2016CharacterRadical} concatenated 
a bi-directional long-short-term memory  (LSTM) network which had learned radical embeddings and character embeddings together, before conditional random field (CRF) learning. Their model achieved better performance on the same task than previous CRF models.
\cite{Wu2019GlyceGF} added glyph vectors of Chinese character representations into Neural Network learning for improved tagging and labelling. 

\begin{figure*}[]
\begin{center}
\centering
\includegraphics*[width=1\textwidth]{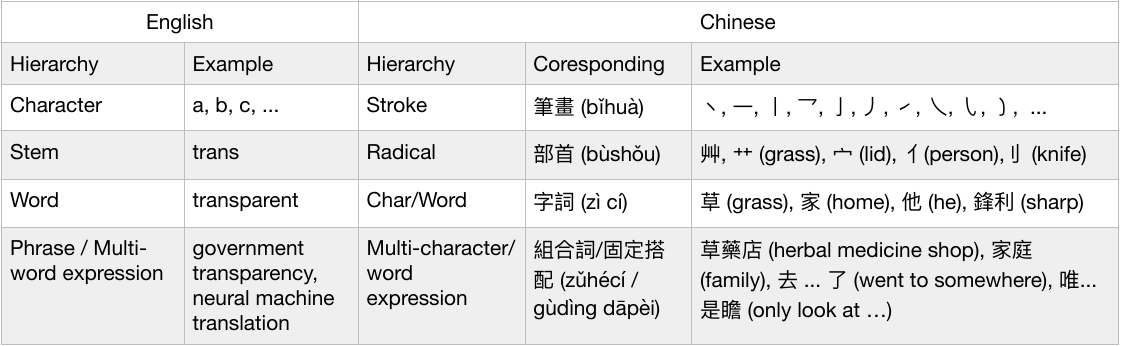}
\caption{A mapping between English and Chinese on compositionality. }
\label{fig:en_zh_linguistic_mapping}
\end{center}
\end{figure*}

In \cite{peng2017radical}, sentence-level sentiment analysis with radical embeddings was
carried out. 
One of the four levels of decomposition, full radical embeddings, achieved the best performance with a convolutional neural network (CNN) when compared to
other traditional classification models such as Naive Bayes (NB).  The authors carried out a matching comparison between English and Chinese in composition, in which they aligned English characters with Chinese radicals. However, we believe a better matching for English characters would be the Chinese strokes while the English stems could better match Chinese radicals from a linguistic point of view. 
The reason for this is that while  English characters and  Chinese strokes do not carry any specific meaning independently unless we group them together in a certain order, the English stem and the Chinese radical carry the meanings of words that they are part of. For example, the English stem ``trans" indicates some knowledge (across) from the word ``transparent", and the Chinese radical ``艹" (cǎozì tóu) indicates some knowledge (grass) from the character ``草" (Cǎo, meaning grass), as explained in
\cite{HanKuang2018NMT}. There are more examples of this in the last section of this paper (Section \ref{intro}). The suggested mapping composition between English and Chinese is shown in Figure~\ref{fig:en_zh_linguistic_mapping}. 

In similar work, \cite{Ke2017radicalRNN} applied radical embeddings with CNNs and recurrent neural networks (RNNs) for sentiment analysis of text in the Chinese and Japanese languages.
Their radical embeddings model achieved similar performance compared to
character and word embeddings, but with the advantage of a much smaller vocabulary. 

Recently, \cite{Zheng2021DecomposeFA} carried out an investigation into the generation of  Chinese definitions using a formation rule called ``modifier-head'' which was inspired by the decomposition of Chinese words into meaningful characters. They proposed the DeFT model which is based on BiLSTM structures enhanced by fusing different word formation features, e.g. ``桃花 (peach-blossom)'' is formed by the characters ``桃 (peach)'' and ``花 (flower)'', and their model proved to be effective in comparison to other work.
Similarly, 
\cite{wu-etal-2021-mect} applied a fuse model using the different granularities of Chinese lexicons, but using Transformer structures for NER tasks. The authors went one step deeper into radicals, i.e. the sub-character information. Their proposed MECD model achieved robust performances on several public data sets including Weibo, Resume, and Ontonotes 4.0.

\subsection{Investigating MWEs in MT}
\label{relatedwork_mwe4mt}

We now look at 
related work  investigating the treatment of MWEs in MT. 
%
MWEs are of interest in many NLP tasks due to their idiosyncratic nature \cite{Sag2002MWE,constant-etal-2017-survey}. 
MWEs can be present in many different linguistic categories including idioms, metaphors, customised phrases, named entities, and (semi-/) fixed expressions, amongst others. 
Examples of this include \emph{kick the bucket} or \emph{apple of my eye}s for idiom and metaphor; \emph{Washington, D.C.}, \emph{National Academy of Science}, and \emph{Donald Trump} for location, institutional and personal names 
respectively. All of these are  fixed expressions. Knowledge of MWE boundaries and their meaning acquisition is very important for many NLP tasks, including MT and information extraction. 

The integration of MWEs into MT generally falls into two categories. The first of these is combining extracted bilingual MWEs into a training corpus to enhance model learning as additional features. The second is using MWEs as a linguistically motivated component to understand and change the machine learning structure itself. 
Of these two 
approaches, the first category has been more popular in the past. For instance, \cite{Ren2009mwe,Bouamor2012IdentifyingBM,rikters2017mwe,han-etal-2020-multimwe} designed a range of methods to extract bilingual MWEs for Chinese $\leftrightarrow$ English, French $\leftrightarrow$ English, German $\rightarrow$ English, and English $\rightarrow$ Czech/Latvian translation respectively for statistical as well as for neural MT models. 
\cite{lambert2005mwe}, on the other hand, grouped MWEs as one token prior to SMT training, while \cite{Li2016neuralname} added 
one chunk layer to a neural network focusing on named entities above the word sequence layers to guide the NMT learning representation. 
Very recent work by \cite{dankers-etal-2022-transformer} tried to interpret pre-trained Transformer models on their capability of learning idioms within the hidden state attentions using English and  seven other European languages.
However, none of this work has 
investigated character-decomposed radical- and stroke-level representations within Chinese MWEs in combination with different decomposition levels of MWEs. 

In this section (\ref{relatedwork_section}), we presented related work on lexical granularity for the MT task, Chinese character decomposition in NLP models, and MWE investigations in MT. These inspired us to propose our own model design which we do in the next section on exploring different levels of Chinese character decomposition into NMT, how these decompositions affect MWE translation, and what their influence is on BiMWE-enhanced MT models.

\section{Model Designs}
\label{section_model_idea}

In this section, we present our computational model designs aimed at exploring word meaning representations, preservation, and Multiword Expression (MWE) interpretations in Neural Machine Translation (NMT). We do this by varying the input units and granularity settings, specifically focusing on the Chinese-English language pair. These experiments will be conducted under constrained training conditions in terms of computation, which means we will limit the size of the training data

Our 
pilot-study model seeks to 
exploit
Chinese character radicals as a semantic feature  to enhance 
character and word-level NMT learning in a recurrent neural network (RNN)-based bidirectional attention structure model, which we 
call  $Radical\_Sem$ (with \textit{Sem} taken from semantics). 

The output of the pilot study will lead to two follow-up experimental investigations. The first set of model variants 
enables us to investigate 
the capability of decomposed character pieces to
represent the original word sequence meaning, with 
decomposition at
different levels from shallow to deep, and to examine 
the translation quality of MWEs which we refer to as 
$Decompose\_Rep$ (where \textit{Rep} is for representation). 
Considering that the Transformer learning structure, as introduced in \cite{google2017attention}, has supplanted many traditional RNN models in Machine Translation (MT) and other Natural Language Processing (NLP) domains since our pilot study phase, we have integrated a Transformer structure into our model which we call $Decompose\_Rep$.

The continuous model 
supports investigation into how bilingual MWE terms perform in a high-performance and rich-resource setting, and its interaction with Model $Decompose\_Rep$, i.e., the decomposition model.
MT researchers have claimed that automatically extracted bilingual MWE pairs can enhance the MT model learning in less-resourced language pairs and low MT-quality but high-resource scenarios. We called this model 
$BiMWE\_Term$.
Both the two follow-on models use a state-of-the-art Transformer \cite{google2017attention} based NMT structure for baseline and for model variations.

\subsection{Pilot Study: $Radical\_Sem$}
\label{subsec_modelI_idea}

The NMT structure in this model is from \cite{kuang-etal-2018-attentionNMT,HanKuang2018NMT} where 
the encoder has a bidirectional RNN structure to encode a source sentence and repeatedly generates hidden vectors over the source sentence.
The decoder is also an RNN structure that predicts the next word given the context vector, the hidden state of the decoder and the previous already-generated word.
This structure follows the guideline of the attention-based NMT of the ``Deep Learning for MT (dl4mt)'' tutorial \footnote{Available at \url{github.com/nyu-dl/dl4mt-tutorial/tree/master/} session2}, which enhances the attention model by feeding the previous word to it.
The model structure in the pilot study is an elaboration of the NMT structure used in \cite{costa-jussa-fonollosa-2016-character}, which explores sub-word level embeddings for the language pair German-English. 

\begin{figure}[h]
\centering
\includegraphics*[width=0.5\textwidth]{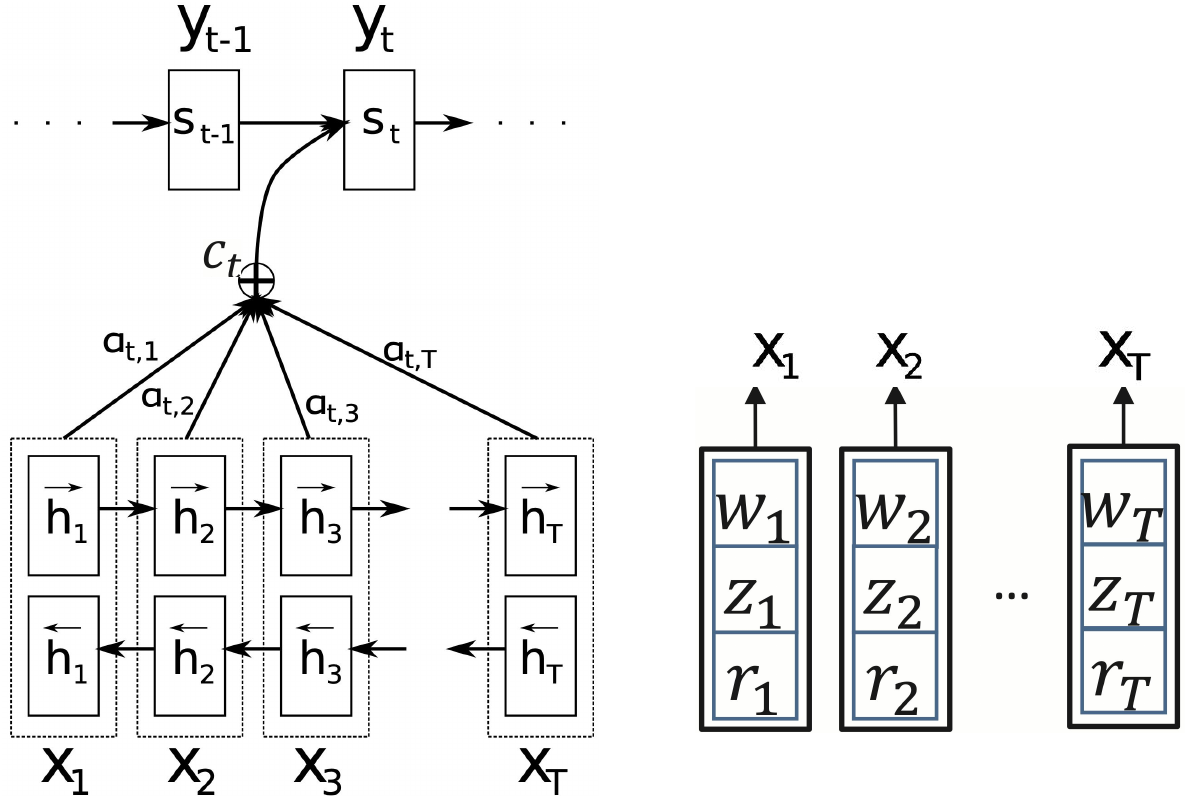}
\caption{BiRNN based NMT architecture \cite{DBLP:journals/corr/BahdanauCB14} using multi-embeddings of (word, character, radical).}
\label{fig:w_c_r_rnn_input}
\end{figure}

In comparison to related work by \cite{chen-etal-2018-combining}, where the authors combined word and character-level knowledge, we explore more diverse input combinations, including word+character+radical represented as triple (word, character, radical) or w+c+r, word+radical represented as two-tuple (word, character) or w+r, character+radical represented as two-tuple (word, radical) or c+r, word+character represented as two-tuple (word, character) or w+c as shown in  Figure~\ref{fig:w_c_r_rnn_input}. We use $w$, $z$ (from 字, zi), and $r$ to represent word, character, and radical respectively as in \cite{HanKuang2018NMT} \footnote{The reason we do not use `c' for the character representation is that we already used c for `context' in this figure and it is a common use for this purpose.}. The detailed experimental settings for our 
investigation of Model-I are indicated in Table~\ref{tab:model_setting_radical2mt}. For this, we use the
radical extraction toolkit 
from HanZiJS \footnote{HanziJS is available at \url{github.com/nieldlr/Hanzi}}.

\begin{table*}[h]
    \centering
\begin{tabular}{ l|c |  c } \toprule
  Investigating Models & Interpretation/Inputs & Indicator \\ \midrule
  BiRNN Baseline & Word sequence & w \\ 
  Variation-I & triple (Word, Character, Radical) & w+c+r \\
  Variation-II & two-tuple (Word, Character) & w+c \\
  Variation-III & two-tuple (Word, Radical) & w+r \\
  Variation-IV & two-tuple (Character, Radical) & c+r \\ \bottomrule
\end{tabular}
\caption{Settings for models investigating semantic NMT with radical enhancement features.}
\label{tab:model_setting_radical2mt}
\end{table*}

\begin{figure*}[h]
\begin{center}
\centering
\includegraphics*[width=.55\textwidth]{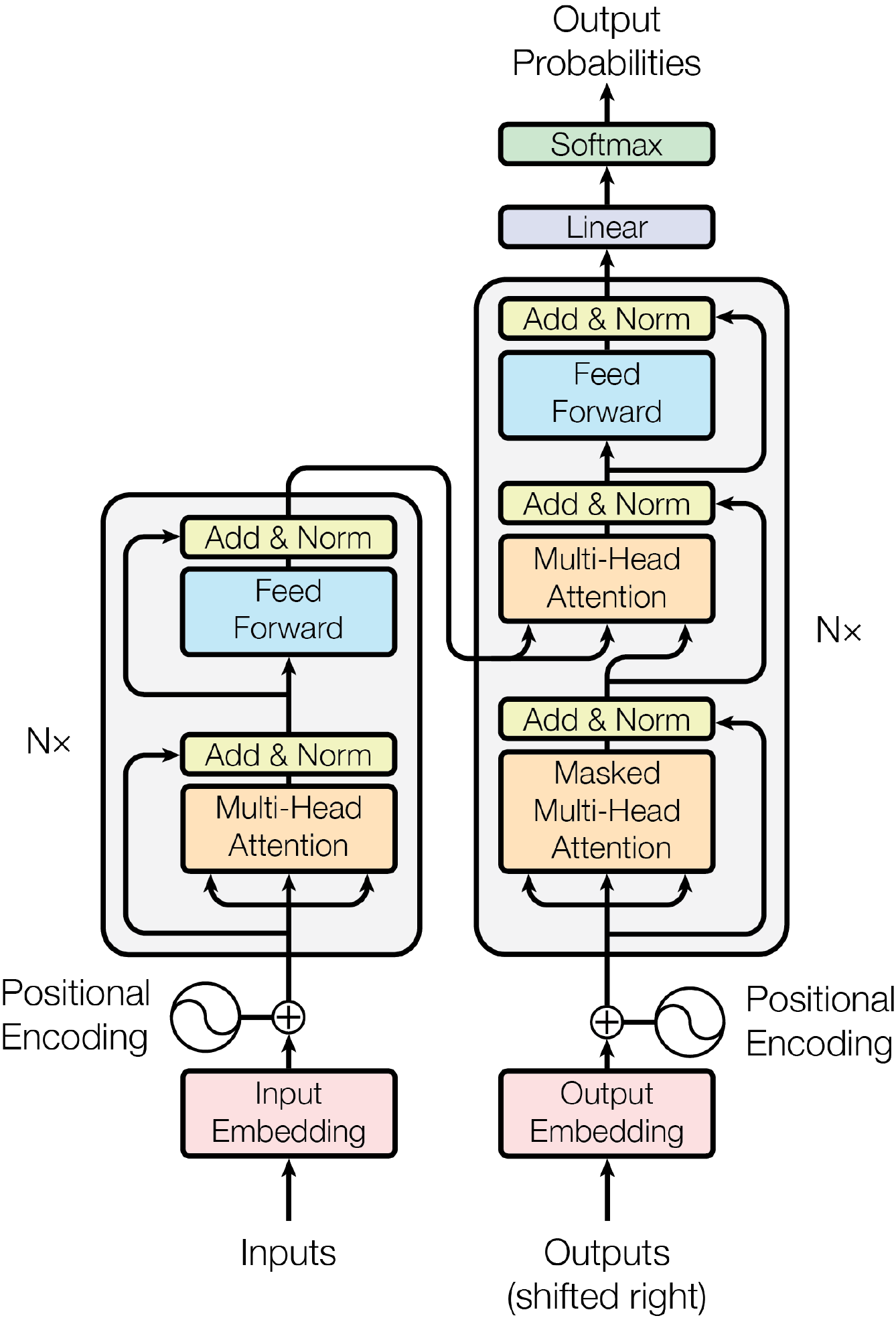}
\caption{Several Attention based Transformer NMT structures \cite{google2017attention}. }
\label{fig:transformer_fig}
\end{center}
\end{figure*}

\subsection{Model: $Decompose\_Rep$ 
}
\label{subsec_modelII_idea}

In the $Decompose\_Rep$ model, following  recent findings by \cite{HanJonesSmeatonBolzoni2021decomposition4mt_MWE} we 
investigate 
how different levels of decomposition of Chinese characters represent the meaning of the original word sequence, in other words where the original word sequences are replaced with word and character pieces with
different degrees of decomposition,
from shallow to deep, with differing granularity. We will use \textsc{rxd}s to indicate the decomposition model in the next section on different degrees, e.g. ``\textsc{rxd}1'', ``\textsc{rxd}2'', and ``\textsc{rxd}3'' for decomposition level-1/2/3 respectively. 
Instead of the BiRNN structure used in the pilot study, we used the state-of-the-art several-attention mechanism Transformer structure designed in \cite{google2017attention} and shown in Figure \ref{fig:transformer_fig}. This
includes source encoder side attention, target decoder side attention, and cross attention between encoder and decoder. For this investigation, we use the open-source THUMT toolkit 
from  Tsinghua University 
\cite{thumt2017}.

To extract different levels of Chinese character decomposition, we use the CHISE project open-source Chinese character dictionary, which 
includes the character structural information and character decomposition pieces\footnote{CHISE: CHaracter Information Service Environment project \url{http://www.chise.org/}}. 
This contains 88,940 Chinese characters from Chinese, Japanese, and Korean (CJK) scripts with Unified ideographs.

Characters can have a single sequence decomposition pattern or up to four possible decomposed representations.
This arises since the characters are from different resources, such as Chinese Hanzi (汉字) used in Chinese Mainland, Hong Kong, and Taiwan which is indicated as G, H, and T respectively, Hanja from Korean (K), ChuNom from Vietnamese (V) and
Kanji from Japanese (J).
\footnote{\textit{ref.} \url{standards.iso.org/ittf/PubliclyAvailableStandards/} for  Universal Coded Character Set (10646:2017).} 

\begin{figure*}[]
\begin{center}
\centering
\includegraphics*[width=1\textwidth]{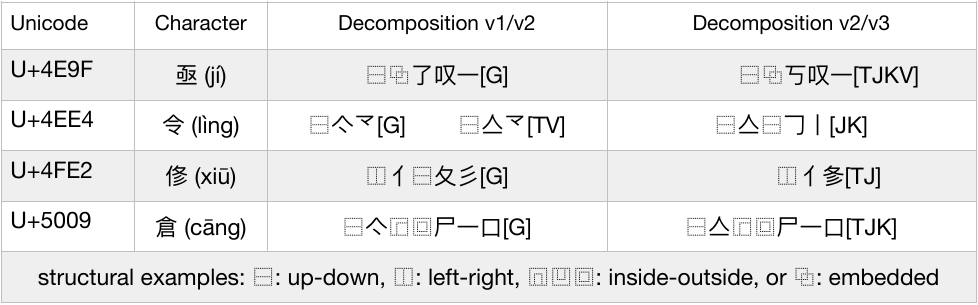}
\caption{Structural construction patterns from example characters offered in IDS. }
\label{fig:ids_examples}
\end{center}
\end{figure*}

Because of these different sources and the changes of languages with societal movement, the appearances of the characters and stroke orders have resulted in certain degrees of variation, for instance  (了 vs. 丂) in the first character example and 
(亽 vs. 亼) in the second and fourth character examples in Figure~\ref{fig:ids_examples}, even though they all have the same root from Chinese Hanzi (汉字, written script from the Han Dynasty). 



In Figure~\ref{fig:ids_examples}, we listed four character examples of which the second character has three alternative decomposition sequences while others have two. In the character decomposition example, the gray patterns indicate the organisational structures of Chinese characters, which can be \textit{up-down, left-right, inside-outside,} or \textit{embedded}, and more. In our experiments, to obtain a level-\textit{L} decomposition of Chinese characters, we run our script through the IDS dictionary \textit{L} times and each time we replace the newly generated smaller-sized characters into their corresponding decomposition sequences recursively.

In situations where there are two or more decomposition styles of a character, we select the Chinese mainland decomposition standard (indicated as G) for the decomposition sequence extraction, since the training corpus we use for experiments corresponds best to the simplified Chinese scripts used in mainland China\footnote{For different data sets with different type of characters, e.g. traditional Chinese, we suggest to use corresponding different standards ``H” (from Hong Kong) or ``T” (from Taiwan) in the character decomposition step. Our character decomposition extraction source code is openly available at \url{https://github.com/poethan/MWE4MT}, so researchers can easily set different parameters for their own experiments, e.g. replacing G with H/T/K(Korean)/V(Vietnamese)/J(Japanese) depending on their data-set deployed.}. 

In the deployed IDS dictionary, wherever the decomposition indicator G is not available for some characters, we choose the first decomposition pattern for our experiments, with detailed explanations as below.
For instance, the character ``叕 (U+53D5, zhuó)'' has two different decomposition choices ``⿱双双'' and ``⿰㕛㕛'', and we use the first ``⿱双双''; for ``並 (U+4E26)'', we use	``⿱䒑业'' instead of ``⿱丷亚''; and for ``串 (U+4E32)'', we use ``⿻吕丨'' instead of ``⿻中口''.
Firstly, from manual examination, we think the first pattern does make more sense in some explainable cases. 
For instance, the character 叕 (zhuó, meaning things repeatedly showing up, mostly negatively) is from the classical Chinese “圣人之思脩，愚人之思叕 (shèng rén zhī sī xiū, yú rén zhī sī zhuó) meaning ``holy admired persons or saints think far and deep; fool folks think all the short things again and again / short-sighted''.

The first decomposition style produces two ``双 (shuāng, meaning double)”, literally indicating ``double and double (replication)”, while the second decomposition style gives two characters 㕛 (yǒu, meaning friend) which is a positive meaning. Thus the first decomposition is preferred.
The other example is the character 串 (chuàn, meaning to use a string to connect things together). The first decomposition generates 吕 (lǚ, a pictographic character, the original character of ridge 脊 (jǐ), which is the shape of two spines 脊骨 (jǐ gǔ) connected) and 丨(pictographic symbol) which is closer to the meaning of using a string to connect things. The second decomposition contains 中(zhōng) and 口 (kǒu), of which 中 have many different common meanings including ``China”, ``middle”, etc. So the first decomposition style is preferred.

Secondly, this phenomenon inspired us to carry out multiple-degree decompositions so this kind of structural ambiguity will be further resolved when a deeper level of decomposition is applied. For instance, the next level of decomposition will convert both medium characters ``㕛 (yǒu)'' and ``双 (shuāng)'' into ``又又 (yòu, again and again)'' (\textit{ref}: U+355B	㕛 $\rightarrow$	⿱又又). This is also one of our motivations to examine the performances from different decomposition degrees.

Again, Figure \ref{fig:decompose_degree_qiaoliang} in Section~\ref{intro} displays the example decomposition degrees of characters ``橋(qiáo, bridge)'' and ``樑(liáng, beam)''.

\subsection{Model: $BiMWE\_Term$}
\label{subsec_modelIII_idea}
In Model $BiMWE\_Term$, we investigate the extraction of bilingual MWE terms from the training corpus and apply these
to the model training stage as augmented data. In addition, we also investigate the effect of decomposed representations of MWEs. The integration of BiMWEs has been shown to be 
very useful in
low-resource settings, e.g. 
\cite{Ren2009mwe,Bouamor2012IdentifyingBM} on the language pairs 
Chinese $\leftrightarrow$ English and French $\leftrightarrow$ English in SMT, and for the high-resource situation 
\cite{rikters2017mwe} on Western language pairs English $\rightarrow$ Czech/Latvian (en~$\rightarrow$~cs/lv) for
NMT, 
see
Section~\ref{relatedwork_mwe4mt}. We observe that even though the NMT systems 
\cite{rikters2017mwe} used 
large 
language resources, e.g. 4.5 million sentences for en~$\rightarrow$~lv and 49 million sentences for en~$\rightarrow$~cs, the baseline translation models do not give 
high-quality performance in terms of 
BLEU scores, i.e. 13.71 and 11.29 for en~$\rightarrow$~cs and en~$\rightarrow$~lv respectively. This
leaves significant scope for improvement of 
translation systems,
since most high-quality translation systems often yield a 20+ BLEU score. 

\begin{figure*}[htb]
\centering
\includegraphics*[width=1\textwidth]{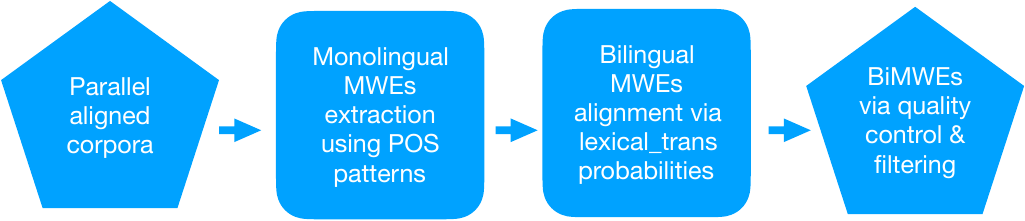}
\caption{Framework for bilingual MWEs (BiMWEs) extraction.}
\label{fig:MultiMWE_workflow}
\end{figure*}

In our
investigation, we adopt the bilingual MWE extraction framework from \cite{rikters2017mwe} for a high-performance MT situation i.e. zh~$\rightarrow$~en using a
state-of-the-art transformer-based structure as our baseline system, as in Model $Decompose\_Rep$. Furthermore, we explore 
word and character decomposed forms of representations of MWEs.
The bilingual MWEs (BiMWEs) are extracted using the workflow by \cite{han-etal-2020-multimwe}, shown in Figure~\ref{fig:MultiMWE_workflow}.

\begin{figure}[ht]
\centering
\includegraphics*[width=.5\textwidth]{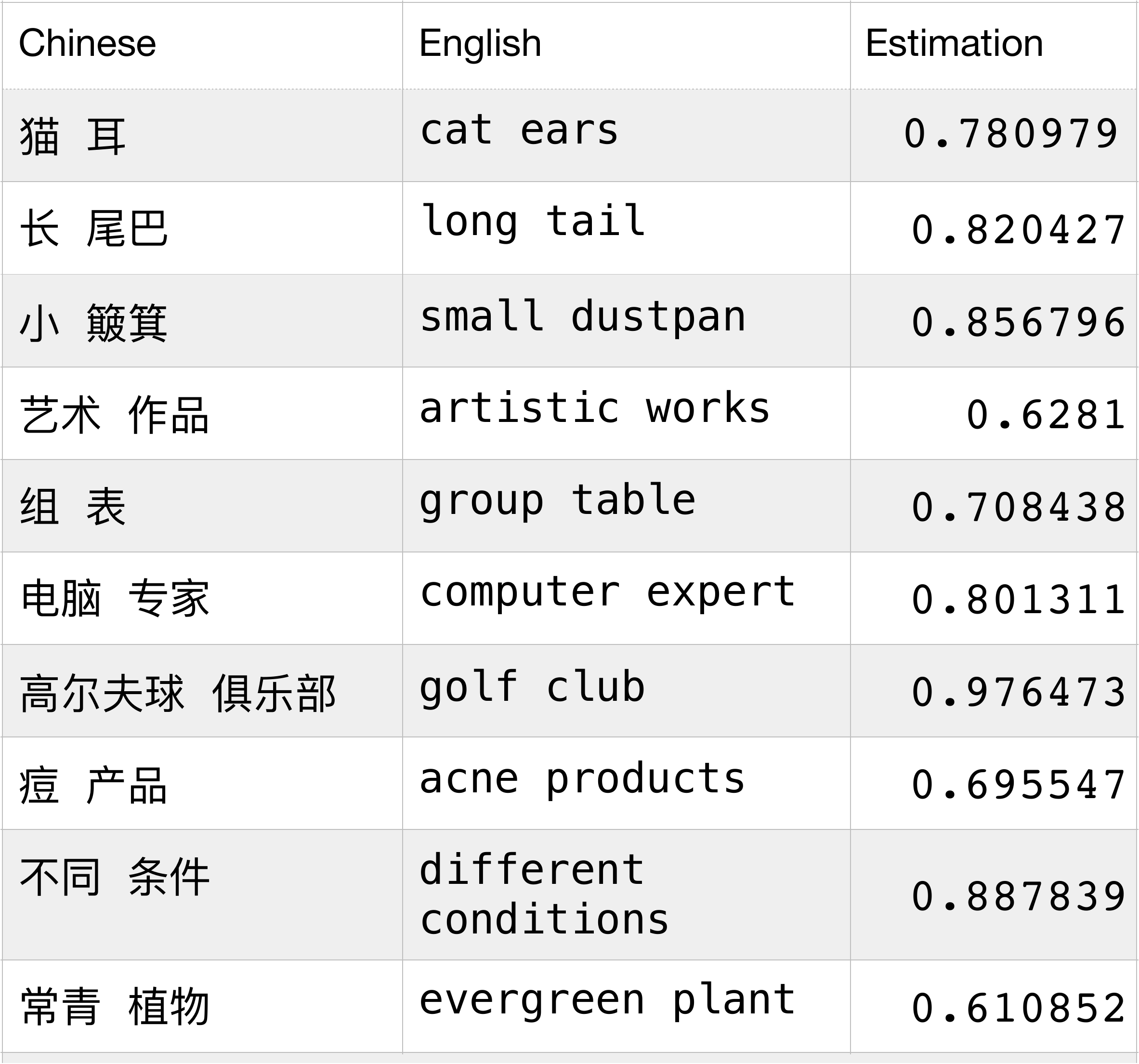}
\caption{Automatically extracted Chinese~$\leftrightarrow$~English BiMWEs before pruning.}
\label{fig:extracted_zh_en_mwe_sample}
\end{figure}

We first use Treetagger \cite{schmid1994treetagger} 
to attach Part-of-Speech (POS) tags and lemma information to English and Chinese sentences.
Subsequently, we design Chinese POS patterns to extract MWEs by mapping from the English patterns used by \cite{rikters2017mwe,han-etal-2020-multimwe} and the manual addition of 
some new tags that indicate fixed expressions, idioms, and named entities including personal names, place names, and organisation names.
We then extract 
English and Chinese monolingual MWEs using the MWEtoolkit \cite{ramisch2015book_mwetoolkit} by deploying the POS patterns we designed\footnote{MWEtoolkit is a pattern-based monolingual MWE extraction tool based on manually defined POS patterns, e.g. adj+adj+noun, noun+noun, verb+adv, etc. \textit{ref.} \url{https://gitlab.com/mwetoolkit/mwetoolkit3/}}. This is reflected in the second block of Figure~\ref{fig:MultiMWE_workflow} (BiMWE workflow).
Next, we deploy the MWE conversion tools and MPaligner \cite{rikters2017mwe,Pinnis2013mp_aligner} to extract bilingual MWE candidates.
We apply filtering to remove low-quality bilingual MWE pairs using a threshold score estimated according to the statistical translation scores of the phrases. This threshold is set to 0.85 in all experiments. 
Finally, we decompose the Chinese MWEs (from bilingual pairs) into radical and stroke sequences to different degrees using the system developed in the Model $Decompose\_Rep$, and add these new bilingual pairs back into the training corpus as augmented data\footnote{All experimental tools are open-source and available at \url{https://github.com/poethan/MWE4MT}}. 
Some zh~$\leftrightarrow$~en example BiMWEs extracted using this framework before pruning (0.85 threshold) are shown in Figure~\ref{fig:extracted_zh_en_mwe_sample}. In the figure, we can see some very good examples of BiMWEs, especially those 
that have high alignment scores, e.g. ``golf club'' alignment to ``高尔夫球 \, 俱乐部 (Gāo ěr fū qiú jù lè bù, or 高爾夫球 \, 俱樂部 in traditional Chinese characters)'' with a score of 0.98.


In summary, in this section, we have introduced our three main model designs for experimental investigation, which include a BiRNN structure NMT with attention included for semantic feature radical testing, a Transformer NMT structure for testing the different levels of Chinese character decompositions, and also using Transformer structure for BiMWE enhanced NMT learning with decomposition characters.
In the next section, we will present details of the performance of each one.
  
\section{Experimental Settings and Evaluations}
\label{model_eval_section}
In this section, we introduce the experimental settings for each of the three models and the corresponding evaluations which are carried out. Because the Model $Decompose\_Rep$ and Model $BiMWE\_Term$ use the same NMT structure and training corpus, we combine the description and analysis of these settings. 

\subsection{Pilot Study Evaluation: $Radical\_Sem$}
\label{subsec_radical4mt_eval}
For the pilot study $Radical\_Sem$ training corpus, 1.25 million parallel sentences are deployed covering 80.9 million and 86.4 million Chinese and English words respectively. The corpus is mostly from LDC Chinese-English parallel data, i.e. LDC2002E18, LDC2003E07, LDC2003E14, LDC2004T07, LDC2004T08, and LDC2005T06, as described in \cite{HanKuang2018NMT}. The NIST MT challenge datasets NIST06 and NIST08 are used as development data and test data respectively, of which up to 4 multiple references are available for the NIST08 testing set.

The BLEU metric \cite{Papineni02bleu:a} is used for model tuning on development data. The maximum sentence length for both source and target languages is set to  50. The sizes of vocabularies are set to be 30k words for both source and target. In the pilot study setting, we replace both side rare words using ``UNK'' (representing unknown), while in our follow-up model investigations via $Decompose\_Rep$ and $BiMWE\_Term$, we will keep
the rare words as part of a different strategy. 
Statistics show the vocabularies covered around (97.7\%, 99.3\%) of the corpora (NIST06, NIST08). 
In this model, the embedding dimensions for (word, character, radical) are all set to 620, and the (word, character, radical) vocabulary size is set to (30k, 2.5k, 1k) according to empirical knowledge.

%

\subsubsection{Broad Evaluation Metrics}

Much 
published work examines
the insufficiency of the BLEU metric for evaluation \cite{BanerjeeLavie2005METEOR,han2012lepor,freitag-etal-2020-bleu,erofeev-etal-2021-cushlepor}.  
%
Being aware of such valuable work, we seek to evaluate our model in a broader evaluation model setting from several state-of-the-art metrics that show better performances than BLEU in the conventional WMT shared task. We choose  hLEPOR \cite{han2013language,han2014lepor}, CharacTER \cite{wang2016character}, and BEER \cite{stanojevic-simaan:2014:W14-33}  in addition to the standard 
 and popular BLEU 
scores \cite{Papineni02bleu:a}.  

BLEU is a popular n-gram-based precision-favoured metric with a geometric mean of each n-gram score, e.g. uni-gram, bi-gram, tri-gram, etc., and the most commonly used parameter is 4-gram. We will report detailed BLEU scores for each model from uni-gram to 4-gram.
The hLEPOR metric incorporates 
multiple factors including sentence length, precision, recall, positional difference, and n-grams at the word level, while CharacTER and BEER carry out character-level evaluation using editing distance, paraphrasing, and syntax information. Among these metrics, higher scores indicate better performances, except for CharacTER which is the opposite. 
All these metrics showed top performances in WMT metrics for shared tasks \cite{machavcek-bojar:2013:WMT,machacek-bojar:2014:W14-33,bojar-EtAl:2016:WMT2,bojar-graham-kamran:2017:WMT} including on the zh$\rightarrow$en language pair\footnote{\url{http://www.statmt.org/wmt17/metrics-task.html}}.

The
hLEPOR metric series have been 
evaluated by other researchers as the best-performing segment-level metrics on par with two others that are not significantly outperformed by any of the rest in a 
comparison of metrics 
\cite{DBLP:conf/naacl/GrahamBM15} using WMT13 data. 

\subsubsection{Evaluations on the Single-reference Development Set NIST06}

The BLEU scores and broader metric scores from the baseline word model and the proposed radical enhanced models using attention-based BiRNN NMT structure on development set NIST06 are shown in Table~\ref{tab:bleu_nist06dev} and  Table~\ref{tab:broader_nist06dev}. 
The numbers in \textbf{bold case}  indicate the best scores and those in \textit{italic case} indicate winning scores over the Baseline.

\begin{table*}[ht]
    \centering
\begin{tabular}{ l|c |c |c | c } \toprule
   & uni-gram & bi-gram & tri-gram & four-gram   \\ \midrule 
  Baseline (BiRNN, W) & 0.7211 & 0.5663 & 0.4480 & 0.3556  \\
  triple (W, C, R) & \textbf{0.7420} & \textbf{0.5783} & \textbf{0.4534} & \textbf{0.3562} \\
  two-tuple (W, C) & \textit{0.7362} &\textit{ 0.5762} & \textit{0.4524} & 0.3555  \\
  two-tuple (W, R) & \textit{0.7346} & \textit{0.5730} & \textit{0.4491} &  0.3529  \\
  two-tuple (C, R) & 0.7089 & 0.5415 & 0.4164 & 0.3219  \\ \bottomrule
\end{tabular}
\caption{Cumulative n-gram BLEU scores on the NIST06 Development Data}
\label{tab:bleu_nist06dev}
\end{table*}

The BLEU scores reflect that the word+character+radical enhanced triple (w, c, r) model produces the highest score across all four-gram categories. Furthermore, just by adding character or radical knowledge the model two-tuple (w, c) and two-tuple (w, r) also outperform the baseline word model in all cases from uni-gram to tri-gram. However, the character+radical two-tuple (c, r) model, i.e., without word-level knowledge, produces a lower score than the baseline word model, which may indicate that for Chinese language translation, the word segmentation or word boundary information is important to modelling learning. 
The first three models in the table 
 verify our assumption for Model $Radical\_Sem$ that when using Chinese radicals as augmented semantic features, the NMT model improves the learning capability on word meanings and translation performance.
Especially, the uni-gram BLEU score of triple (w, c, r) achieves around 2.1 absolute score gain in comparison to the word-based baseline, which indicates the adequacy level of the translation is much improved. This aspect of improvement for the NMT model is very important since there are researchers pointing out that NMT improved the MT fluency level while adequacy level performance often has a drop in comparison to SMT, e.g. \cite{DBLP:journals/corr/TuLLLL16,koehn-knowles-2017-six}. 


\begin{table*}[ht]
    \centering
\begin{tabular}{ l|c  |c |  c } \hline 
   & \multicolumn{3}{c}{Evaluation on single-reference}  \\ \hline 
  Model variations  & ~~hLEPOR~~ & ~~BEER~~ & ~~CharacTER~~ \\ \hline
  Baseline (BiRNN, W)~~ & 0.5890 & 0.5112 & 0.9225  \\
  triple (W, C, R)  & \textit{0.5972} & \textbf{0.5167} & \textbf{0.9169}  \\
  two-tuple (W, C)  & \textbf{0.5988} & \textit{0.5164} & 0.9779  \\
  two-tuple (W, R)  & \textit{0.5942} & \textit{0.5146} &  0.9568 \\
  two-tuple (C, R)  & 0.5779 & 0.4998 & 1.336  \\ \hline 
\end{tabular}
\caption{Evaluation scores from broader metrics using the NIST06 development set.}
    \label{tab:broader_nist06dev}
\end{table*}

The results from broader evaluation metrics on the development set in Table~\ref{tab:broader_nist06dev} indicate that the triple (w, c, r) model wins all three metrics over Baseline, while the two-tuple (w, c) and two-tuple (w, r) win two metrics hLEPOR and BEER over Baseline. In addition, the triple (w, c, r) model achieves two best scores, while two-tuple (w, c) gets one of the best scores on hLEPOR.
Consistently, the two-tuple (c, r) model without using word-level knowledge produces the worst scores across the three metrics.

In Model $Decompose\_Rep$, we will further investigate the granularity of input units, and simplify this aspect by only using the decomposed character pieces with the original word boundary information reserved.


\subsubsection{Evaluations on the Multi-reference Testing Set NIST08}

The evaluation scores using BLEU, and broader metrics on four-reference-based and case-insensitive NIST08 testing corpus are shown in Tables~\ref{tab:bleu_nist08test} and Table~\ref{tab:broader_nist08test} \cite{HanKuang2018NMT}.

The evaluation scores from the BLEU metric on the testing set show similar findings to the development set scores, where the triple (w, c, r) model produces the best scores for all four-gram categories. 
Interestingly, the two-tuple (w, r) model, i.e. word+radical, performs better scores than both the two-tuple (w, c), i.e. word+character, and word baseline model in uni-gram and bi-gram situations, which verifies the assumption of our research that the decomposition of Chinese characters into semantic radical can help model learning as augmented knowledge features.

\begin{table*}[ht]
    \centering
\begin{tabular}{ l|c |c |c | c} \hline 
   & uni-gram & bi-gram & tri-gram & four-gram   \\ \hline 
  Baseline (BiRNN, W) & 0.6451 & 0.4732 & 0.3508 & 0.2630  \\
  triple (W, C, R) & \textbf{0.6609} & \textbf{0.4839} & \textbf{0.3572} & \textbf{0.2655} \\
  two-tuple (W, C) & 0.6391 & 0.4663 & 0.3412 & 0.2527 \\
  two-tuple (W, R) & \textit{0.6474} & \textit{0.4736} & 0.3503 &  0.2607  \\
  two-tuple (C, R) & 0.6378 & 0.4573 & 0.3296 & 0.2410  \\ \hline 
\end{tabular}
\caption{Cumulative n-gram BLEU scores on the NIST08 Testing Data}
    \label{tab:bleu_nist08test}
\end{table*}

\begin{table*}[ht]
    \centering
\begin{tabular}{ l|c |c  |  c } \hline 
   & \multicolumn{3}{c}{4-references based evaluation}  \\ \hline 
  Model variations  & hLEPOR & BEER & CharacTER \\ \hline
  Baseline (BiRNN, W) & 0.5519 & 0.4748 & \textbf{0.9846} \\
  triple (W, C, R) & \textbf{0.5530} & \textbf{0.4778} & 1.3514  \\
  two-tuple (W, C) &  0.5444 & 0.4712 & 1.1416  \\
  two-tuple (W, R) &  0.5458 & 0.4717 &  0.9882 \\
  two-tuple (C, R) &  0.5353 & 0.4634 & 1.1888  \\ \hline 
\end{tabular}
\caption{Evaluation scores from broader metrics using the NIST08 testing set.}
    \label{tab:broader_nist08test}
\end{table*}

From the broader evaluation metrics, the triple (w, c, r) model achieves the two best scores in hLEPOR and BEER, while the word baseline model wins CharacTER.

Overall, the designed model variations using character and radical enhancement are best-performing in most of the evaluations from the development set to the testing set. Among these comparisons, there are both the popular system-level metric BLEU and one of the best-performing segment-level metrics, hLEPOR. 
On the testing set, the word+radical model wins the word+character model on all metrics including BLEU, hLEPOR, BEER, and CharacTER, which indicates that radical knowledge can be a better replacement for character knowledge to boost word-level attention-based BiRNN structure NMT learning.
With this finding, we continue to carry out an empirical investigation in the next section where deeper level decomposition than the radical level will be tested.

\subsection{Model Evaluations: $Decompose\_Rep$ \& $BiMWE\_Term$}

As  discussed in  Section~\ref{relatedwork_granularity4mt}, Section~\ref{relatedwork_zh_decompose4nlp},
and Section~\ref{subsec_modelII_idea},
Model $Decompose\_Rep$ enables further investigation into the granularity of NMT input units and examines the
capability 
of the decomposed word and character pieces to represent the original word sequence (or, character sequence with word boundaries) in 
translation. 
Because both Model $Decompose\_Rep$ and Model $BiMWE\_Term$ use the same corpus and Transformer NMT structure for baseline settings, we carry out the evaluations of these two models simultaneously.

We first investigate translation behaviour for different decomposition levels of Chinese characters, from shallow to deep, as shown in Figure~\ref{fig:decompose_degree_qiaoliang} using radical and stroke sequences. The goal is to reduce the number of unknown and rare words, and low-frequency MWEs. 
Then, to investigate how such decomposition models can affect the BiMWEs enhancement-based NMT learning structure, we apply the decomposition models to the extracted BiMWEs as well, which we described in Section ~\ref{subsec_modelIII_idea}.

In order to achieve the generalisability of our decomposition model, the training corpus we use for the Model-II and Model-III experiments is a five million portion of parallel sentences from the WMT-2018 MT shared task \cite{wmt2018findings}, larger than the corpus size used in the pilot study model $Radical\_Sem$ investigation. 
The development set and testing set are from WMT-2017 
which have 2001 and 2002 zh$\Leftrightarrow$en sentences \cite{wmt2017findings}.
 
In light of the findings from Model $Radical\_Sem$ in the last section that Chinese word boundaries do play a helpful role in model learning and translation performances, we preserve the word boundary information and only replace the original word and character representation with corresponding decomposed word and character pieces for the two further investigation models.

The NMT toolkit we use is THUMT which is a Tensorflow version implementation of the Transformer model. We set up the encoder and decoder to 7+7 layers, with batch size set to 6,250, and 32k BPE operations.

We carry out empirical evaluation using both BLEU metric and crowd-sourced human evaluation using Direct Assessment (DA) \cite{NLE:9961497}. 
Being aware of the very recent research work on MT evaluation by \cite{google2021human_evaluation_TQA} that the automatic BLEU scores ranking has a closer correlation to the crowd-sourced based human evaluation, than to human expert translators' professional evaluations, we also carry out expert validations looking into output examples, especially the translation of MWEs.

\subsubsection{BLEU and Crowd-sourced Based Evaluations}

\begin{table*}[htb]
\begin{center}
\begin{tabular}{lcccccc} \toprule
                & 20k &  100k & 120k & 140k & 160k & 180k \\ 
                \midrule
baseline    & 18.39  & \textbf{21.56} & 21.45         & 21.31 & 21.29 & 21.42 \\  
base+BiMWE &\textbf{ 18.49}           & 21.39          & \textbf{21.67} & \textbf{21.83} & \textbf{21.42} & \textbf{21.86} \\  
\textsc{rxd3}  & 16.48  & 20.75 &20.73 &20.93 &20.98 &21.14 \\ 
\textsc{rxd3}+BiMWE & 17.82  &21.36  &21.50 & 21.31   &\textbf{21.42}& 21.47   \\ 
\textsc{rxd2} &11.84  &13.26 &12.88&13.02   &13.38 &12.86   \\  
\textsc{rxd1}/ideograph &15.52 & 20.67& 20.61&21.26   & 20.76& 21.00 \\  
  \bottomrule
 \end{tabular}
\caption{Evaluation scores using the BLEU metric from 20k to 180k learning steps.
}
\label{tab:zh2en_wmt18_bleu_xk}
\end{center}
\end{table*} 

\begin{table}[!t]
\begin{center}
\centering
\small 
\begin{tabular}{crccc}
\toprule
\multicolumn{1}{c}{Ave. raw} 
     & \multicolumn{1}{c}{Ave. $z$}     
                & $n$   & $N$   & \\
\midrule
73.2 & 0.161    & 1,232 & 1,639 & \textsc{base(word)}    \\
71.6 & 0.125    & 1,262 & 1,659 & \textsc{word+BiMWE}     \\
71.6 & 0.113    & 1,257 & 1,672 & \textsc{rxd1}    \\ 
71.3 & 0.109    & 1,214 & 1,593 & \textsc{rxd3+BiMWE} \\ 
70.2 & 0.073    & 1,260 & 1,626 & \textsc{rxd3}    \\ \hline
53.9 & $-$0.533 & 1,227 & 1,620 & \textsc{rxd2}    \\  
\bottomrule
\end{tabular}
\caption{Crowd-sourced based human evaluation using DA. 
Ave. raw indicates the average raw score from DA; Ave. $z$ is the standardised score per human assessor.  $n$ and $N$ are the number of distinct translations and the number of assessments including repetition. The horizontal line indicates the above models form a cluster without significant difference
while the model below the line is much worse than those above it \cite{HanJonesSmeatonBolzoni2021decomposition4mt_MWE}.}
\label{fig:sys_level_he_score}
\end{center}
\end{table}

\begin{figure*}
\begin{center}
\includegraphics*[width=1\textwidth]{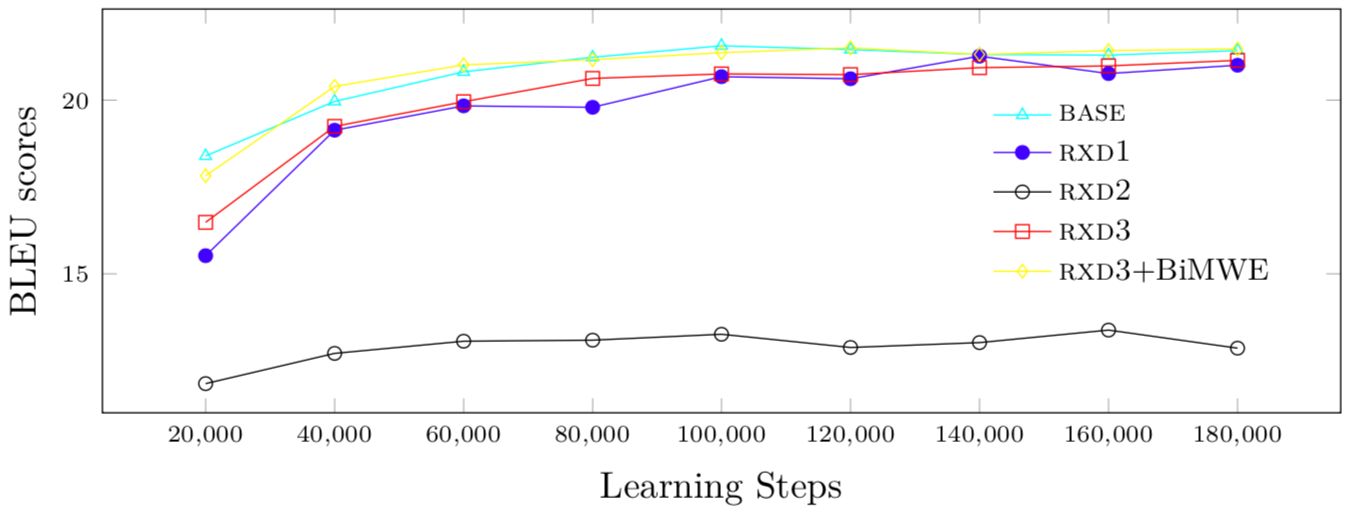}
\caption{Learning curves from 20k to 180k learning steps using BLEU from word based model, decomposition models and BiMWE enhanced decomposition model \cite{HanJonesSmeatonBolzoni2021decomposition4mt_MWE}.}\label{steps}
\end{center}
\end{figure*} 

In this sub-section, we present both BLEU metric scores and crowd-sourced based human evaluation scores following the investigation from \cite{HanJonesSmeatonBolzoni2021decomposition4mt_MWE}. In the next sub-section, we will present our expert evaluations on sample outputs.

We use the Direct Assessment (DA) method that has been deployed in the WMT shared task for Amazon Mechanical Turk (MTurk) based crowd-sourced human evaluation \footnote{\url{https://www.mturk.com}}.
For DA, we use the model translation outputs from 180K learning steps, using the quality control measures of \cite{NLE:9961497}.
The evaluation scores from BLEU and DA are shown in Table~\ref{tab:zh2en_wmt18_bleu_xk} and Table~\ref{fig:sys_level_he_score} respectively, where we display the BiMWE enhanced word-level NMT model base+BiMWE and decomposition models \textsc{rxd}s. Since the decomposition level-3 model (\textsc{rxd}3) 
produces relatively higher scores compared to the other two decomposition models, we also add the model performance of \textsc{rxd}3+BiMWE. The decomposition level-1 model, i.e. \textsc{rxd}1, is the level of decomposition when the semantic (radical) and phonetic parts of the Chinese characters are initially separated, which is called the \textit{ideograph} model in  \cite{zhang-komachi-2018-neural}. We also draw a learning curve using BLEU scores in Figure~\ref{steps}. 

BLEU scores across 180K learning steps show that in most cases, BiMWE enhanced word-based NMT model wins the best performance, except that Baseline is best-performing with 100K learning steps and \textsc{rxd}3+BiMWE is on par with base+BiMWE with 160K learning steps. Thus, the BLEU metric indicates that BiMWE-enhanced knowledge can boost both a word-based model and decomposition model in a constrained setting (5M sentences), even though it is already a high-performance language translation condition, e.g. 20+ BLEU scores.

However, the crowd-sourced based DA scores as shown in Table~\ref{fig:sys_level_he_score} do not reflect exactly the same findings as from the BLEU metric. All five research models achieve high performance which is on par without statistical significance, forming one cluster. This means that the ranking among these five models can change  with greater feeding of sentences for evaluation but they will not move to a different cluster. 
However, the DA result does indicate the decomposition level-II (\textsc{rxd2}) model performs at a lower level (cluster) than others (at confidence level  $p$< 0.05).

Another beneficial aspect we learned from this comparison is that the decomposition models generate much fewer system parameters than word-based models that the NMT systems need to learn. This indicates a possibility of reducing the computational complexity of the overall neural structure. 
For instance, the total trainable variable sizes of the word-based baseline model and decomposition level-III model are (89,456,896 vs. 80,288,000), which means the \textsc{rxd3} model reduces 10.25\% of the overall parameters to be more space efficient\footnote{All data collected in the crowd-sourced human evaluation is available at the same repository \url{https://github.com/poethan/MWE4MT}.}.

In summary, while BLEU scores show better performances achieved by augmented models using BiMWEs, e.g. base+BiMWE model, in comparison to a baseline model, the crowd-sourced DA assessment does not reflect significant differences between these models. 
One possible explanation for this could be that the BiMWE-augmented NMT can be helpful for low-resource (e.g. < 1 million training sample sentences) language pair settings, or low-performance (< 20 BLEU scores) high-resource settings (e.g. \cite{rikters2017mwe}), but for language pairs with both high-resources and high-performance, the BiMWE augmentation strategy may not be influential such as in a setting of 5+ million training sentences and 7+7 layers of Transformer-based NMT in the News domain.
To further investigate this aspect, we carry out an expert validation in the next sub-section.

\subsubsection{Expert Validations of Examples with MWEs}

Due to the inconsistency between the BLEU score and crowd-sourced-based DA assessment, as well as the uncertainty surrounding the confidence level of crowd-sourced human work, e.g., human professional translator based validations on historical WMT data by \cite{google2021human_evaluation_TQA}, an extra step of human-expert validation of the system's outputs is regularly carried out, with particular attention paid to the translation of MWEs. Two native Chinese speakers are involved in the expert-based validation, one Master's graduate and one PhD graduate both of whom work in the NLP field and speak fluent English. We have one of them first carry out their expert evaluation and annotation, then the second one conducts quality checking. 

\begin{figure*}[]
\begin{center}
\centering
\includegraphics*[width=1\textwidth]{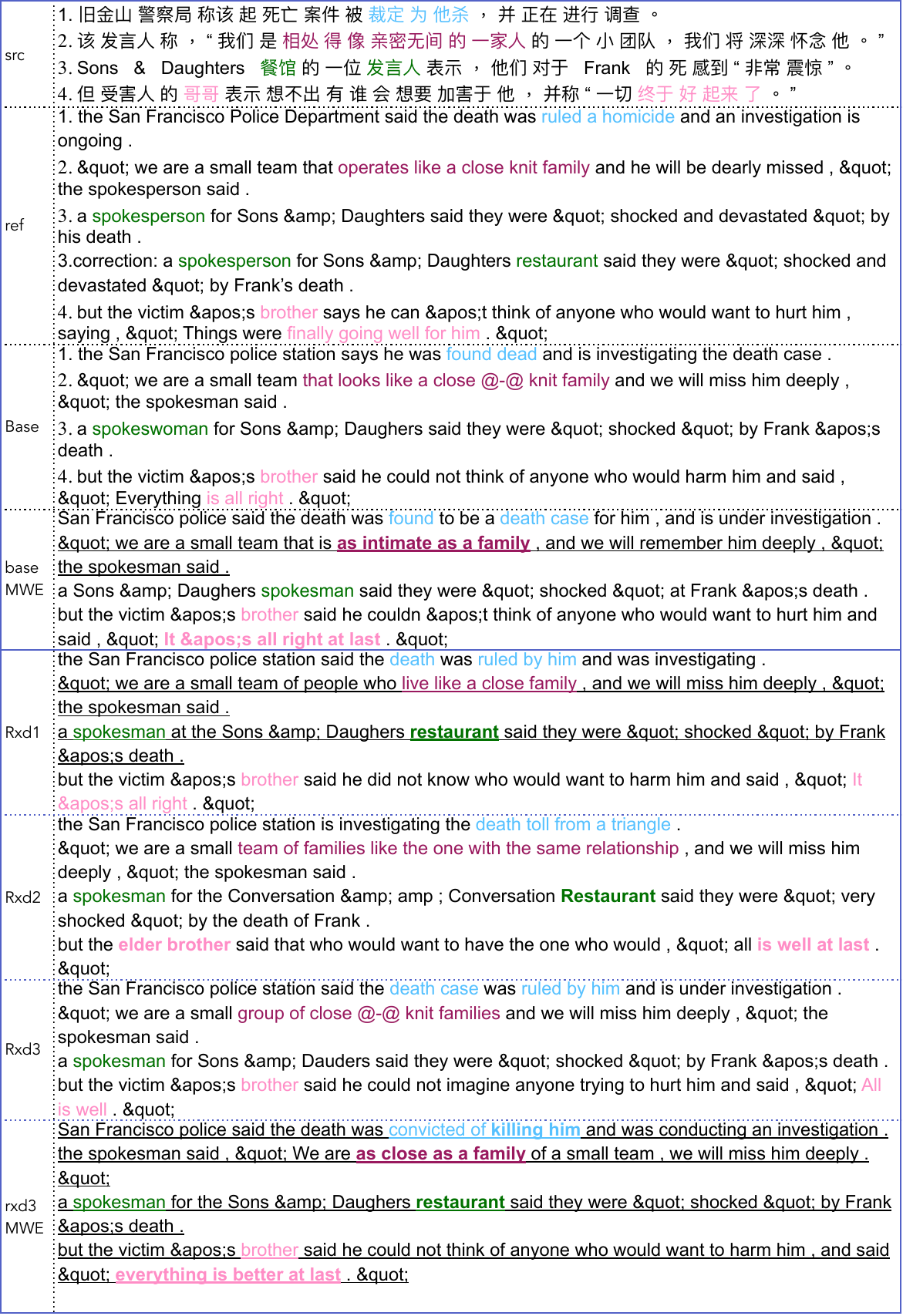}
\caption{MT output comparisons from models including Base, baseMWE, \textsc{rxd}s and \textsc{rxd}3MWE using sample sentences from src, translated from 180k learning steps.
\textsc{rxd}s indicate the decomposition models (level 1, 2, 3). MWE indicates the BiMWE enhanced model. Bold case and underline indicate best performances.}
\label{fig:MToutput_rxds_MWEs_vs_base_180k}
\end{center}
\end{figure*}

For the expert-based validation set, we used the heading part of the test set of around 30 sentences.
We asked the evaluators to mark all  errors in the MT outputs, similar to the study in \cite{popovic-2020-informative} and especially to pay attention to MWE-related errors. 
Out of these sentences, Figure~\ref{fig:MToutput_rxds_MWEs_vs_base_180k} includes the worked-out sample set of sentences of source, reference, and MT outputs. We use an underline to indicate the best overall translation at the sentence level and use bold case to indicate the correctly translated terms/MWEs highlighted in coloured font.

The sample sentences for experts' validation are translation outputs from 180k learning steps, i.e., the most trained of the translation models. We use these sample sentences to acquire an in-depth view of the errors made by different models, i.e., word sequence-based models including \textit{base} and BiMWE enhanced word sequence models \textit{base+BiMWE}, and decomposition models including \textit{\textsc{rxd}1}; \textit{\textsc{rxd}2}; \textit{\textsc{rxd}3}; and BiMWE enhanced decomposition model \textit{\textsc{rxd}3+BiMWE}.

Findings from the expert validation show that the BiMWE-enhanced models and decomposition models generate overall better translation outputs at sentence-level than word sequence-based strong baseline models despite having very similar BLEU scores, with the exception of the \textsc{rxd}2 model. This tells us that expert validation can reveal interesting findings that cannot be revealed by the BLEU score or crowd-sourced human evaluations. We explain this in detail below.

Sentence 1, ``裁定为他杀\, (ruled as a homicide)'' is translated as ``found dead'' by the baseline model, which overlooks the meaning of the original MWE ``他杀\,(other+kill $\rightarrow$ killed by others $\rightarrow$ homicide)''. The reason for that is  the word ``他\,(him/he)'', which has an ambiguous meaning in this context. Instead of the commonly-used meaning of ``him or he'', in this context, ``他\,'' is shortened from ``他人\, (others)'', so the MWE term ``他杀'' indicates ``killed (杀) by \textit{somebody else}''. 
All the models fail to correctly translate this ambiguous MWE term, except for the \textsc{rxd}3+BiMWE model, which successfully translates it into a meaning equivalent phrase ``convicted of killing him'', but this cannot be reflected by n-gram matching BLEU scores due to the use of different words from the reference translation ``ruled a homicide''.

In sentence 2, the phrase ``相处 得 像 亲密无间 的 一家人\, (operates like a close-knit family)'' is translated by the baseline as ``that looks like a close-knit family'', which is not very accurate. The translations by base+BiMWE, \textsc{rxd}1 and \textsc{rxd}3+BiMWE have better meaning equivalent outputs to interpret the MWE ``相处 得\, (operates like, instead of \textit{looks like})'' using terms ``as intimate as'', ``live like'', and ``as close as''.

For sentence 3, the reference translation in the data offered by the WMT   is actually plainly wrong, and we corrected it in the figure presentation as ``a spokesperson for Sons and Daughters restaurant said they were shocked and devastated by Frank's death''. The baseline model surprisingly translated the MWE ``发言人\,(spokesperson)'' into a female gender ``spokeswoman'', even though other models produce ``spokesman''. The baseline model also drops out the MWE term ``餐馆\,(restaurant)'', while the decomposition models \textsc{rxd}1, \textsc{rxd}2 and \textsc{rxd}3+BiMWE all translate this term correctly.

For sentence 4, the MWE ``哥哥\,(elder brother)'' is only translated correctly by \textsc{rxd}2, and all other models translate it into ``brother''. The phrase ``终于 好 起来 了\, (finally going well (for him))'' is translated as ``is all right'' by the baseline model, which drops the meaning that ``things were not good in the beginning'' which is reflected by MWEs ``终于\,(finally)'' and ``好 起来\, (getting better)''. This phrase is better translated by base+BiMWE, \textsc{rxd}2, and \textsc{rxd}3+BiMWE by ``is all right at last'', ``is well at last'', and ``is better at last''.

Overall, we can see the baseline model performs more poorly compared to the other variation models we proposed in all four sample sentences. Furthermore, in addition to improving MWE translations, our proposed models also have better performance with regard to terminology translations, for instance, on the terms ``homicide \textit{vs} dead” from sentence-1, and ``spokesperson \textit{vs} spokesman” in sentence-3. Terminologies can appear as single-word or multi-words, and their translation in different domains such as medical and legal texts has been an open issue in MT research for a long period  including both SMT and NMT, especially in a low-resource scenario \cite{xiong_etal_2016_topicTermMT,haque-etal-2019-term_MT}. 
We also note that the decomposition models can help with better term translation accuracy. Potentially this can  be further applied to other ideographic language translations such as Japanese and Korean. In this work, we designed three variation models focusing on the investigation of how Chinese character decomposition can help low-frequency words, phrases, and especially MWE translations. In the future, we intend to carry out separate experiments on how decomposition models can improve term translation with quantitative and qualitative analysis in selected domains, e.g. using medical or legal data.


We also discovered that the \textsc{rxd2} model can actually  translate some MWE terms better than the baseline model even though its overall sentence-level translation quality is lower. For instance, \textsc{rxd2} translates ``哥哥\,(gē gē)'' correctly into ``elder brother'' which is the only correct translation from all competitors, and it also translates ``餐馆\,(cān guǎn)'' into ``restaurant'' correctly unlike word-based baseline and base+BiMWE models.

The \textsc{rxd3}+BiMWE model is the best-performing model, with the \textsc{rxd1} model being next,  closely followed by base+BiMWE. This indicates that both BiMWEs and character decomposition contribute to translation model learning and that their combination contributes to translation accuracy to a larger degree. This verifies the assumption of our main focus, i.e., if the decomposition of Chinese characters helps the word meaning representation for the MT task, and if it affects the BiMWE integrated NMT learning structure towards more accurate translation of MWEs. 

Our other findings include that the testing corpus needs to be improved, e.g., in sentence 3 where the offered testing reference has two errors: missing ``restaurant" entity translation; and error translation from ``Frank’s" into ``his".
The reference of sentence 4 also has one error of ``brother’’ instead of ``elder brother".
This indicates the issues in the testing suite that WMT (the Annual Workshop of MT) and MT communities have, and the need for higher quality alternative testing-suit creation, e.g. reducing errors and introducing multiple reference semantic equivalent translations.

To obtain a further in-depth understanding of \textsc{rxd2}, decomposition level 2, for which the automatic BLEU score and crowd-sourced human evaluations gave a much lower ranking compared to other models, we looked into the decomposition examples and this led to another insight to share.
Level 1 of decomposition often separates the Chinese characters into a phonetic part and a semantic part, as the example in Figure~\ref{fig:decompose_degree_qiaoliang} of characters qiáo and liáng, and deeper levels (e.g. Level 3) often generates deeper decomposition of characters with closer to stroke-level character pieces with sequence order. However, Level 2 decomposition tends to produce smaller-sized characters that can be independent with new meanings that are not directly related to the root character. For instance, 吞 (tūn) and 冋 (jiōng) are produced in Level 2 of character 橋 (qiáo, bridge), however, they have very different meanings, from bridge, of ``swallow'' and ``the face of a person when he/she does not know if he/she shall laugh or cry by the situation, e.g. a person opens his/her mouth and becomes numb as indicated by the presentation of this character''.

Similarly, 刅\,(chuāng) is produced in Level 2 from the character 樑 (liáng, beam, a structural element to resist loads), however, it means ``cut, wound, trauma'' which is distinct from the meaning of its root character. Level 3 decomposition will further split this smaller-sized character 刅\,(chuāng) into a partial sequence with strokes.

In summary, this section presented the experimental setup and evaluation outputs of our three models, $Radical\_Sem$, $Decompose\_Rep$, and $BiMWE\_Term$. We examined separately how Chinese radicals can be used as semantic features, how different character decomposition levels perform, and how these affect BiMWE augmented NMT learning and translation performances on low-frequency words, phrases, and MWEs.

\section{Discussions and Conclusions}
\label{discussion_conclusions}

Our empirical investigation reported in this paper is motivated by the connections between word meaning representations and multi-word expression acquisitions in an automatic machine translation setting. 
We chose Chinese-English as the investigating language pair since there are large numbers of data resources available as open source, and this will be helpful for further potential research work to be carried out based on ours. The Chinese language has a sophisticated writing script and this leaves much opportunity to explore better representations of words and MWEs in meaning acquisition and translation. 

As argued by  recent work from the MT field, e.g. \cite{L_ubli_2020_human_parity}, the Chinese-English NMT output is still far from reaching real human expert parity. 
To further advance the state-of-the-art NMT systems' performance, we carried out an empirical study using BiRNN and Transformer-based strong NMT learning structures. 
We tackled the MT challenges on low-frequency words, phrases, and MWEs by investigating different levels of Chinese character decompositions in a systematic examination. Little work has been done  in this aspect previously in the MT field, especially focusing on ideographic languages which include Chinese, Japanese, Korean, and many others, even though sub-word modelling has been widely adopted for Western alphabetic language processing \cite{SubwordNMT15Sennrich,kudo-2018-subword}. 
We hope our work will be helpful in shedding light on ideographic language processing regarding word meaning representation, MWEs acquisition and translation, and even beyond. 

Furthermore, we re-examined the automated extraction of BiMWEs and their application as data-augmented learning for NMT in a state-of-the-art transformer-based structure using both character sequence and character decompositions. Our findings show that in a high-performing and high-resource zh~$\rightarrow$~en language setting, the BiMWEs augmented strategy does still give a positive effect in improving translation accuracy, such as on MWEs, for a word-based model as reflected by our experts' validation (base+BiMWE model). However, the decomposed models show much better capability in handling this aspect with the effect of BiMWEs (e.g. the \textsc{rxd}3+BiMWE model)
as a promising alternative solution at least for the Chinese language. This may even lead to further improvement with model refinement in the future.

\section*{Acknowledgements}
LH and GJ thank the ADAPT Centre for Digital Content Technology funded under the SFI Research Centres Programme (Grant 13/RC/2106) and co-funded under the European Regional Development Fund. AS thanks the Insight Centre for Data Analytics funded by Science Foundation Ireland under grant number SFI/12/RC/2289 (Insight Centre). We thank Yvette Graham, Shaohui Kuang, and Paolo Bolzoni for the valuable input and contribution to our previous work that builds the foundation to this one. 


\section{Bibliographical References}\label{sec:reference}

\bibliographystyle{lrec2026-natbib}
\bibliography{lrec2026-example}

@LanguageResource{hantowards,
  title={Towards a resource for multilingual lexicons: an MT assisted and human-in-the-loop multilingual parallel corpus with multi-word expression annotation},
  year = "2025",
  url = "https://github.com/aaronlifenghan/AlphaMWE/releases/tag/corpus",
  organization="GitHub Open-source",
  edition = "AlphaMWE 2.0",
  author={Han, Lifeng and Mohamed, Najet Hadj and Rassem, Malak and Jones, Gareth JF and Smeaton, Alan F and Nenadic, Goran}
}

@inproceedings{hadj-mohamed-etal-2023-alphamwe,
    title = "{A}lpha{MWE}-{A}rabic: {A}rabic Edition of Multilingual Parallel Corpora with Multiword Expression Annotations",
    author = "Hadj Mohamed, Najet  and
      Rassem, Malak  and
      Han, Lifeng  and
      Nenadic, Goran",
    editor = "Mitkov, Ruslan  and
      Angelova, Galia",
    booktitle = "Proceedings of the 14th International Conference on Recent Advances in Natural Language Processing",
    month = sep,
    year = "2023",
    address = "Varna, Bulgaria",
    publisher = "INCOMA Ltd., Shoumen, Bulgaria",
    url = "https://aclanthology.org/2023.ranlp-1.50/",
    pages = "448--457"
}

@phdthesis{han2022investigation,
  title={An investigation into multi-word expressions in machine translation},
  author={Han, Lifeng},
  year={2022},
  school={Dublin City University}
}

@InProceedings{han_gladkoff_metaeval_tutorial2022,
  title={Meta-Evaluation of Translation Evaluation Methods: a systematic up-to-date overview},
  author={Han, Lifeng and Gladkoff, Serge},
  url={https://github.com/poethan/LREC22_MetaEval_Tutorial},
  booktitle ={Tutorial at LREC2022},
  address = {Marseille, France},
  year ={2022}
}

@inproceedings{wu-etal-2021-mect,
    title = "{MECT}: {M}ulti-Metadata Embedding based Cross-Transformer for {C}hinese Named Entity Recognition",
    author = "Wu, Shuang  and
      Song, Xiaoning  and
      Feng, Zhenhua",
    booktitle = "Proceedings of the 59th Annual Meeting of the Association for Computational Linguistics and the 11th International Joint Conference on Natural Language Processing (Volume 1: Long Papers)",
    month = aug,
    year = "2021",
    address = "Online",
    publisher = "Association for Computational Linguistics",
    url = "https://aclanthology.org/2021.acl-long.121",
    doi = "10.18653/v1/2021.acl-long.121",
    pages = "1529--1539",
}

@inproceedings{Zheng2021DecomposeFA,
  title={Decompose, Fuse and Generate: A Formation-Informed Method for Chinese Definition Generation},
  author={Hua Zheng and Damai Dai and Lei Li and Tianyu Liu and Zhifang Sui and Baobao Chang and Yang Liu},
  booktitle={North American Chapter of the Association for Computational Linguistics},
  year={2021},
  url={https://api.semanticscholar.org/CorpusID:235097253}
}

@inproceedings{dankers-etal-2022-transformer,
    title = "Can Transformer be Too Compositional? Analysing Idiom Processing in Neural Machine Translation",
    author = "Dankers, Verna  and
      Lucas, Christopher  and
      Titov, Ivan",
    booktitle = "Proceedings of the 60th Annual Meeting of the Association for Computational Linguistics (Volume 1: Long Papers)",
    month = may,
    year = "2022",
    address = "Dublin, Ireland",
    publisher = "Association for Computational Linguistics",
    url = "https://aclanthology.org/2022.acl-long.252",
    doi = "10.18653/v1/2022.acl-long.252",
    pages = "3608--3626",
}

@inproceedings{popovic-2020-informative,
    title = "Informative Manual Evaluation of Machine Translation Output",
    author = "Popovi{\'c}, Maja",
    booktitle = "Proceedings of the 28th International Conference on Computational Linguistics",
    month = dec,
    year = "2020",
    address = "Barcelona, Spain (Online)",
    publisher = "International Committee on Computational Linguistics",
    url = "https://aclanthology.org/2020.coling-main.444",
    doi = "10.18653/v1/2020.coling-main.444",
    pages = "5059--5069",
}

@inproceedings{haque-etal-2019-term_MT,
    title = "Investigating Terminology Translation in Statistical and Neural Machine Translation: A Case Study on {E}nglish-to-{H}indi and {H}indi-to-{E}nglish",
    author = "Haque, Rejwanul  and
      Hasanuzzaman, Md  and
      Way, Andy",
    booktitle = "Proceedings of the International Conference on Recent Advances in Natural Language Processing ({RANLP} 2019)",
    month = sep,
    year = "2019",
    address = "Varna, Bulgaria",
    publisher = "INCOMA Ltd.",
    url = "https://aclanthology.org/R19-1052",
    doi = "10.26615/978-954-452-056-4_052",
    pages = "437--446",
}

@article{xiong_etal_2016_topicTermMT,
author = {Xiong, Deyi and Meng, Fandong and Liu, Qun},
title = {Topic-Based Term Translation Models for Statistical Machine Translation},
year = {2016},
issue_date = {March 2016},
publisher = {Elsevier Science Publishers Ltd.},
address = {GBR},
volume = {232},
number = {C},
issn = {0004-3702},
url = {https://doi.org/10.1016/j.artint.2015.12.002},
doi = {10.1016/j.artint.2015.12.002},
journal = {Artif. Intell.},
month = {mar},
pages = {54–75},
numpages = {22}
}

@inproceedings{kudo-2018-subword,
    title = "Subword Regularization: Improving Neural Network Translation Models with Multiple Subword Candidates",
    author = "Kudo, Taku",
    booktitle = "Proceedings of the 56th Annual Meeting of the Association for Computational Linguistics (Volume 1: Long Papers)",
    month = jul,
    year = "2018",
    address = "Melbourne, Australia",
    publisher = "Association for Computational Linguistics",
    url = "https://aclanthology.org/P18-1007",
    doi = "10.18653/v1/P18-1007",
    pages = "66--75",
}

@article{L_ubli_2020_human_parity,
   title={A Set of Recommendations for Assessing Human–Machine Parity in Language Translation},
   volume={67},
   ISSN={1076-9757},
   DOI={10.1613/jair.1.11371},
   journal={Journal of Artificial Intelligence Research},
   publisher={AI Access Foundation},
   author={Läubli, Samuel and Castilho, Sheila and Neubig, Graham and Sennrich, Rico and Shen, Qinlan and Toral, Antonio},
   year={2020},
   month={Mar}
}

@inproceedings{liu-ng-2012-character,
    title = "Character-Level Machine Translation Evaluation for Languages with Ambiguous Word Boundaries",
    author = "Liu, Chang  and
      Ng, Hwee Tou",
    booktitle = "Proceedings of the 50th Annual Meeting of the Association for Computational Linguistics (Volume 1: Long Papers)",
    month = jul,
    year = "2012",
    address = "Jeju Island, Korea",
    publisher = "Association for Computational Linguistics",
    url = "https://aclanthology.org/P12-1097",
    pages = "921--929",
}

@inproceedings{freitag-etal-2020-bleu,
    title = "{BLEU} might be Guilty but References are not Innocent",
    author = "Freitag, Markus  and
      Grangier, David  and
      Caswell, Isaac",
    booktitle = "Proceedings of the 2020 Conference on Empirical Methods in Natural Language Processing ({EMNLP})",
    month = nov,
    year = "2020",
    address = "Online",
    publisher = "Association for Computational Linguistics",
    url = "https://aclanthology.org/2020.emnlp-main.5",
    doi = "10.18653/v1/2020.emnlp-main.5",
    pages = "61--71",
}

@article{Salesky2020OptimizingSG_nmt,
  title={Optimizing segmentation granularity for neural machine translation},
  author={Elizabeth Salesky and Andrew Runge and Alex Coda and Jan Niehues and Graham Neubig},
  journal={Machine Translation},
  year={2020},
  volume={34},
  pages={41-59}
}

@inproceedings{BanerjeeLavie2005METEOR,
	Author = {Satanjeev Banerjee and Alon Lavie},
	Booktitle = {Proceedings of the Annual Meeting of the Association for Computational Linguistics},
	Title = {{METEOR: An Automatic Metric for MT Evaluation with Improved Correlation with Human Judgments}},
	Year = {2005}}

@inproceedings{kuang-etal-2018-attentionNMT,
    title = "Attention Focusing for Neural Machine Translation by Bridging Source and Target Embeddings",
    author = "Kuang, Shaohui  and
      Li, Junhui  and
      Branco, Ant{\'o}nio  and
      Luo, Weihua  and
      Xiong, Deyi",
    booktitle = "Proceedings of the 56th Annual Meeting of the Association for Computational Linguistics (Volume 1: Long Papers)",
    month = jul,
    year = "2018",
    address = "Melbourne, Australia",
    publisher = "Association for Computational Linguistics",
    url = "https://www.aclweb.org/anthology/P18-1164",
    doi = "10.18653/v1/P18-1164",
    pages = "1767--1776",
}

@INPROCEEDINGS{Park_Zhao2020_Korean_nmt,  
author={Park, Jeonghyeok and Zhao, Hai},  booktitle={2020 International Conference on {A}sian Language Processing (IALP)},   title={Korean Neural Machine Translation Using Hierarchical Word Structure},   year={2020},  volume={},  number={},  pages={294-298},  doi={10.1109/IALP51396.2020.9310510}}

@inproceedings{stratos-2017-sub_korean_nlp,
    title = "A Sub-Character Architecture for {K}orean Language Processing",
    author = "Stratos, Karl",
    booktitle = "Proceedings of the 2017 Conference on Empirical Methods in Natural Language Processing",
    month = sep,
    year = "2017",
    address = "Copenhagen, Denmark",
    publisher = "Association for Computational Linguistics",
    url = "https://aclanthology.org/D17-1075",
    doi = "10.18653/v1/D17-1075",
    pages = "721--726",
}

@inproceedings{beinborn-etal-2013-cognate,
    title = "Cognate Production using Character-based Machine Translation",
    author = "Beinborn, Lisa  and
      Zesch, Torsten  and
      Gurevych, Iryna",
    booktitle = "Proceedings of the Sixth International Joint Conference on Natural Language Processing",
    month = oct,
    year = "2013",
    address = "Nagoya, Japan",
    publisher = "{A}sian Federation of Natural Language Processing",
    url = "https://aclanthology.org/I13-1112",
    pages = "883--891",
}

@inproceedings{costa-jussa-fonollosa-2016-character,
    title = "Character-based Neural Machine Translation",
    author = "Costa-Juss{\`a}, Marta R.  and
      Fonollosa, Jos{\'e} A. R.",
    booktitle = "Proceedings of the 54th Annual Meeting of the Association for Computational Linguistics (Volume 2: Short Papers)",
    month = aug,
    year = "2016",
    address = "Berlin, Germany",
    publisher = "Association for Computational Linguistics",
    url = "https://aclanthology.org/P16-2058",
    doi = "10.18653/v1/P16-2058",
    pages = "357--361",
}

@inproceedings{tiedemann-2012-character,
    title = "Character-Based Pivot Translation for Under-Resourced Languages and Domains",
    author = {Tiedemann, J{\"o}rg},
    booktitle = "Proceedings of the 13th Conference of the {E}uropean Chapter of the Association for Computational Linguistics",
    month = apr,
    year = "2012",
    address = "Avignon, France",
    publisher = "Association for Computational Linguistics",
    url = "https://aclanthology.org/E12-1015",
    pages = "141--151",
}

@inproceedings{costa-jussa-etal-2017-byte,
    title = "Byte-based Neural Machine Translation",
    author = "Costa-juss{\`a}, Marta R.  and
      Escolano, Carlos  and
      Fonollosa, Jos{\'e} A. R.",
    booktitle = "Proceedings of the First Workshop on Subword and Character Level Models in {NLP}",
    month = sep,
    year = "2017",
    address = "Copenhagen, Denmark",
    publisher = "Association for Computational Linguistics",
    url = "https://aclanthology.org/W17-4123",
    doi = "10.18653/v1/W17-4123",
    pages = "154--158",
}

@inproceedings{dai-yamaguchi-2019-compact,
    title = "Compact and Robust Models for {J}apanese-{E}nglish Character-level Machine Translation",
    author = "Dai, Jinan  and
      Yamaguchi, Kazunori",
    booktitle = "Proceedings of the 6th Workshop on {A}sian Translation",
    month = nov,
    year = "2019",
    address = "Hong Kong, China",
    publisher = "Association for Computational Linguistics",
    url = "https://aclanthology.org/D19-5202",
    doi = "10.18653/v1/D19-5202",
    pages = "36--44",
}

@inproceedings{chen-etal-2018-combining,
    title = "Combining Character and Word Information in Neural Machine Translation Using a Multi-Level Attention",
    author = "Chen, Huadong  and
      Huang, Shujian  and
      Chiang, David  and
      Dai, Xinyu  and
      Chen, Jiajun",
    booktitle = "Proceedings of the 2018 Conference of the {N}orth {A}merican Chapter of the Association for Computational Linguistics: Human Language Technologies, Volume 1 (Long Papers)",
    month = jun,
    year = "2018",
    address = "New Orleans, Louisiana",
    publisher = "Association for Computational Linguistics",
    url = "https://aclanthology.org/N18-1116",
    doi = "10.18653/v1/N18-1116",
    pages = "1284--1293",
}

@inproceedings{han-etal-2020-alphamwe,
    title = "{A}lpha{MWE}: Construction of Multilingual Parallel Corpora with {MWE} Annotations",
    author = "Han, Lifeng  and
      Jones, Gareth  and
      Smeaton, Alan F.",
    booktitle = "Proceedings of the Joint Workshop on Multiword Expressions and Electronic Lexicons",
    month = dec,
    year = "2020",
    address = "online",
    publisher = "Association for Computational Linguistics",
    url = "https://www.aclweb.org/anthology/2020.mwe-1.6",
    pages = "44--57",
}

@ARTICLE{google2021human_evaluation_TQA,
       author = {{Freitag}, Markus and {Foster}, George and {Grangier}, David and {Ratnakar}, Viresh and {Tan}, Qijun and {Macherey}, Wolfgang},
        title = "{Experts, Errors, and Context: A Large-Scale Study of Human Evaluation for Machine Translation}",
      journal = {arXiv e-prints},
         year = 2021,
        month = apr,
          eid = {arXiv:2104.14478},
        pages = {arXiv:2104.14478},
archivePrefix = {arXiv},
       eprint = {2104.14478},
 primaryClass = {cs.CL},
       adsurl = {https://ui.adsabs.harvard.edu/abs/2021arXiv210414478F}
}

@article{NLE:9961497,
  author = {Graham, Yvette and Baldwin, Timothy and Moffat, Alistair and Zobel, Justin},
  title = {Can machine translation systems be evaluated by the crowd alone},
  journal = {Natural Language Engineering},
  volume = {FirstView},
  month = {1},
  year = {2016},
  issn = {1469-8110},
  pages = {1--28},
  numpages = {28},
  doi = {10.1017/S1351324915000339},
  URL = {http://journals.cambridge.org/article_S1351324915000339},
}

@inproceedings{HanJonesSmeatonBolzoni2021decomposition4mt_MWE,
    title = "{C}hinese Character Decomposition for Neural {MT} with Multi-Word Expressions",
    author = "Han, Lifeng  and
      Jones, Gareth  and
      Smeaton, Alan F. and
      Bolzoni, Paolo",
    booktitle = "Proceedings of the 23rd Nordic Conference on Computational Linguistics (NoDaLiDa)",
    month = may # " 31--2 " # jun,
    year = "2021",
    address = "Reykjavik, Iceland (Online)",
    publisher = {Link{\"o}ping University Electronic Press, Sweden},
    url = "https://aclanthology.org/2021.nodalida-main.35",
    pages = "336--344",
}

@inproceedings{Dong2016CharacterRadical,
  title={Character-Based {LSTM-CRF} with Radical-Level Features for Chinese Named Entity Recognition},
  author={Chuanhai Dong and Jiajun Zhang and Chengqing Zong and Masanori Hattori and Hui Di},
  booktitle={NLPCC/ICCPOL},
  year={2016}
}

@inproceedings{Cao2018cw2vecLC,
  title={cw2vec: Learning Chinese Word Embeddings with Stroke n-gram Information},
  author={Shaosheng Cao and Wei Lu and Jun Zhou and Xiaolong Li},
  booktitle={Thirty-Second AAAI Conference on Artificial Intelligence},
  year={2018}
}

@article{Wu2019GlyceGF,
  title={Glyce: Glyph-vectors for {C}hinese Character Representations},
  author={Wei Wu and Yuxian Meng and Qinghong Han and Muyu Li and Xiaoya Li and Jie Mei and Ping Nie and Xiaofei Sun and Jiwei Li},
  journal={Annual Conference
               on Neural Information Processing Systems},
  year={2019}
}

@inproceedings{peng2017radical,
	author = {Haiyun Peng and Erik Cambria and Xiaomei Zou},
	title = {Radical-Based Hierarchical Embeddings for Chinese Sentiment Analysis at Sentence Level},
	booktitle = {Florida Artificial Intelligence Research Society Conference},
	year = {2017},

	url = {https://aaai.org/ocs/index.php/FLAIRS/FLAIRS17/paper/view/15460}
}

@article{Ke2017radicalRNN,
  author    = {Yuanzhi Ke and
               Masafumi Hagiwara},
  title     = {Radical-level Ideograph Encoder for RNN-based Sentiment Analysis of
               {Chinese and Japanese}},
  journal   = {CoRR},
  volume    = {abs/1708.03312},
  year      = {2017},
  url       = {http://arxiv.org/abs/1708.03312},
  archivePrefix = {arXiv},
  eprint    = {1708.03312},
  bibsource = {dblp computer science bibliography, https://dblp.org}
}

@inproceedings{google2017attention,
  author={Ashish Vaswani and Noam Shazeer and Niki Parmar and Jakob Uszkoreit and Llion Jones and Aidan N. Gomez and Lukasz Kaiser and Illia Polosukhin},
  title={Attention is All you Need},
  year={2017},
  cdate={1483228800000},
  pages={6000-6010},
  booktitle={ Conference on Neural Information Processing System},
}

@inproceedings{rikters2017mwe,
	author = {Rikters, Matīss and Bojar, Ond\v{r}ej},
	booktitle={Proceedings of the 16th Machine Translation Summit (MT Summit 2017)},
	title = {{Paying Attention to Multi-Word Expressions in Neural Machine Translation}},
	address={Nagoya, Japan},
	year = {2017}
}

@article{cho2014encoder-decoder,
  author    = {KyungHyun Cho and
               Bart van Merrienboer and
               Dzmitry Bahdanau and
               Yoshua Bengio},
  title     = {On the Properties of Neural Machine Translation: Encoder-Decoder Approaches},
  journal   = {CoRR},
  volume    = {abs/1409.1259},
  year      = {2014},
  url       = {http://arxiv.org/abs/1409.1259},
  archivePrefix = {arXiv},
  eprint    = {1409.1259},
  bibsource = {dblp computer science bibliography, https://dblp.org}
}

@article{constant-etal-2017-survey,
    title = "{S}urvey: Multiword Expression Processing: A {S}urvey",
    author = {Constant, Mathieu  and
      Eryi{\v{g}}it, G{\"u}l{\c{s}}en  and
      Monti, Johanna  and
      Plas, Lonneke van der  and
      Ramisch, Carlos  and
      Rosner, Michael  and
      Todirascu, Amalia},
    journal = "Computational Linguistics",
    volume = "43",
    number = "4",
    month = dec,
    year = "2017",
    url = "https://www.aclweb.org/anthology/J17-4005",
    doi = "10.1162/COLI_a_00302",
    pages = "837--892",
}

@article{ZhangMatsumoto18radical,
  author    = {Jinyi Zhang and
               Tadahiro Matsumoto},
  title     = {Improving Character-level {Japanese-Chinese} Neural Machine Translation
               with Radicals as an Additional Input Feature},
  journal   = {CoRR},
  volume    = {abs/1805.02937},
  year      = {2018},
  url       = {http://arxiv.org/abs/1805.02937},
  archivePrefix = {arXiv},
  eprint    = {1805.02937},
  bibsource = {dblp computer science bibliography, https://dblp.org}
}

@inproceedings{zhang-komachi-2018-neural,
    title = "Neural Machine Translation of Logographic Language Using Sub-character Level Information",
    author = "Zhang, Longtu  and
      Komachi, Mamoru",
    booktitle = "Proceedings of the Third Conference on Machine Translation: Research Papers",
    month = oct,
    year = "2018",
    address = "Brussels, Belgium",
    publisher = "Association for Computational Linguistics",
    url = "https://www.aclweb.org/anthology/W18-6303",
    doi = "10.18653/v1/W18-6303",
    pages = "17--25",
}

@InProceedings{Li2016neuralname,
author="Li, Xiaoqing
and Yan, Jinghui
and Zhang, Jiajun
and Zong, Chengqing",
editor="Chen, Jiajun
and Zhang, Jiajun",
title="Neural Name Translation Improves Neural Machine Translation",
booktitle="Machine Translation",
year="2019",
publisher="Springer Singapore",
address="Singapore",
pages="93--100",
isbn="978-981-13-3083-4"
}

@incollection{Savarytv2018vMWE,
	author={Agata Savary and  Marie Candito and  Verginica Barbu Mititelu and  Eduard Bejček and  Fabienne Cap and  Slavomír Čéplö and  Silvio Ricardo Cordeiro and  Gülşen Eryiğit and  Voula Giouli and  Maarten van Gompel and  Yaakov HaCohen-Kerner and  Jolanta Kovalevskaitė and  Simon Krek and  Chaya Liebeskind and  Johanna Monti and  Carla Parra Escartín and  Lonneke van der Plas and  Behrang QasemiZadeh and  Carlos Ramisch and  Federico Sangati and  Ivelina Stoyanova  and  Veronika Vincze  },
	title={{PARSEME multilingual corpus of verbal multiword expressions}},
	booktitle={{Multiword expressions at length and in depth{: Extended papers from the MWE 2017 workshop}}},
    publisher={Language Science Press.},
	Address={Berlin},
	year={2018},
	pages={87--147},
	doi={10.5281/zenodo.1469555},
	}

@inproceedings{Papineni02bleu:a,
	Author = {Kishore Papineni and Salim Roukos and Todd Ward and Wei Jing Zhu},
	Booktitle = {Proceedings of Annual Meeting of the Association for Computational Linguistics},
	Title = {{BLEU}: a method for automatic evaluation of machine translation},
	Year = {2002}}

@book{Koehn2010,
	Author = {Philipp Koehn},
	Publisher = {Cambridge University Press},
	Title = {Statistical Machine Translation},
	Year = {2010}}

@inproceedings{DBLP:conf/naacl/GrahamBM15,
  author    = {Yvette Graham and
               Timothy Baldwin and
               Nitika Mathur},
  title     = {Accurate Evaluation of Segment-level Machine Translation Metrics},
  booktitle = {{NAACL} {HLT} 2015, The 2015 Conference of the {North American} Chapter
               of the Association for Computational Linguistics: Human Language Technologies,
               {Denver, Colorado, USA, May 31 - June 5, 2015}},
  pages     = {1183--1191},
  year      = {2015},
  url       = {http://aclweb.org/anthology/N/N15/N15-1124.pdf},
  bibsource = {dblp computer science bibliography, http://dblp.org}
}

@article{thumt2017,
  title={{THUMT}: An Open Source Toolkit for Neural Machine Translation},
  author={Jiacheng Zhang and Yanzhuo Ding and Shiqi Shen and Yong Cheng and Maosong Sun and Huan-Bo Luan and Yang Liu},
  journal={ArXiv},
  year={2017},
  volume={abs/1706.06415}
}

@inproceedings{DBLP:journals/corr/TuLLLL16,
  author    = {Zhaopeng Tu and
               Zhengdong Lu and
               Yang Liu and
               Xiaohua Liu and
               Hang Li},
  title     = {Modeling Coverage for Neural Machine Translation},
  booktitle = {Proceedings of the 54th Annual Meeting of the Association for Computational
               Linguistics, {ACL} 2016, {A}ugust 7-12, 2016, Berlin, Germany, Volume
               1: Long Papers},
  year      = {2016},
  url       = {https://www.aclweb.org/anthology/P16-1008/},
  bibsource = {dblp computer science bibliography, https://dblp.org}
}

@article{brown-etal-1993-mathematics,
    title = "The Mathematics of Statistical Machine Translation: Parameter Estimation",
    author = "Brown, Peter F.  and
      Della Pietra, Stephen A.  and
      Della Pietra, Vincent J.  and
      Mercer, Robert L.",
    journal = "Computational Linguistics",
    volume = "19",
    number = "2",
    year = "1993",
    url = "https://www.aclweb.org/anthology/J93-2003",
    pages = "263--311",
}

@inproceedings{han2013language,
  title={Language-independent Model for Machine Translation Evaluation with Reinforced Factors},
  author={Han, Lifeng and Wong, Derek F. and Chao, Lidia S. and He, Liangye and Lu, Yi and Xing, Junwen and Zeng, Xiaodong},
  booktitle={Machine Translation Summit XIV},
  pages={215--222},
  year={2013},
  organization={International Association for Machine Translation}
}

@inproceedings{wang2016character,
  title={CharacTer: Translation Edit Rate on Character Level.},
  author={Wang, Weiyue and Peter, Jan-Thorsten and Rosendahl, Hendrik and Ney, Hermann},
  booktitle={{WMT}},
  pages={505--510},
  year={2016}
}

@InProceedings{bojar-EtAl:2016:WMT2,
  author    = {Bojar, Ond\v{r}ej  and  Graham, Yvette  and  Kamran, Amir  and  Stanojevi\'{c}, Milo\v{s}},
  title     = {Results of the WMT16 Metrics Shared Task},
  booktitle = {Proceedings of the First Conference on Machine Translation},
  month     = {{A}ugust},
  year      = {2016},
  address   = {Berlin, Germany},
  publisher = {Association for Computational Linguistics},
  pages     = {199--231},
  url       = {http://www.aclweb.org/anthology/W/W16/W16-2302}
}

@InProceedings{stanojevic-simaan:2014:W14-33,
  author    = {Stanojevi\'{c}, Milo\v{s} and  Sima'an, Khalil},
  title     = {BEER: BEtter Evaluation as Ranking},
  booktitle = {Proceedings of the Ninth Workshop on Statistical Machine Translation},
  month     = {June},
  year      = {2014},
  address   = {Baltimore, Maryland, USA},
  publisher = {Association for Computational Linguistics},
  pages     = {414--419},
  url       = {http://www.aclweb.org/anthology/W/W14/W14-3354}
}

@InProceedings{machavcek-bojar:2013:WMT,
  author    = {Mach\'{a}\v{c}ek, Matou\v{s}  and  Bojar, Ond\v{r}ej},
  title     = {Results of the {WMT13} Metrics Shared Task},
  booktitle = {Proceedings of the Eighth Workshop on Statistical Machine Translation},
  month     = {{A}ugust},
  year      = {2013},
  address   = {Sofia, Bulgaria},
  publisher = {Association for Computational Linguistics},
  pages     = {45--51},
  url       = {http://www.aclweb.org/anthology/W13-2202}
}

@InProceedings{machacek-bojar:2014:W14-33,
  author    = {Machacek, Matous  and  Bojar, Ondrej},
  title     = {Results of the {WMT14} Metrics Shared Task},
  booktitle = {Proceedings of the Ninth Workshop on Statistical Machine Translation},
  month     = {June},
  year      = {2014},
  address   = {Baltimore, Maryland, USA},
  publisher = {Association for Computational Linguistics},
  pages     = {293--301},
  url       = {http://www.aclweb.org/anthology/W/W14/W14-3336}
}

@InProceedings{bojar-graham-kamran:2017:WMT,
  author    = {Bojar, Ond\v{r}ej and Graham, Yvette and Kamran, Amir},
  title     = {Results of the {WMT}17 Metrics Shared Task},
  booktitle = {Proceedings of the Second Conference on Machine Translation, Volume 2: Shared Tasks Papers},
  month     = {September},
  year      = {2017},
  address   = {Copenhagen, Denmark},
  publisher = {Association for Computational Linguistics}
}

@inproceedings{HanKuang2018NMT,
 author = {Lifeng Han and Shaohui Kuang},
 title = {Incorporating {C}hinese Radicals Into Neural Machine Translation: Deeper Than Character Level},
 booktitle = {Proceedings of {ESSLLI-2018}},
 year = {2018},
 month = {{A}ugust},
 location = {Sofia, Bulgaria},
 pages = {54--65},
 numpages = {12},
 publisher = {Association for Logic, Language and Information ({FoLLI})},
}

@inproceedings{DBLP:journals/corr/BahdanauCB14,
  author       = {Dzmitry Bahdanau and
                  Kyunghyun Cho and
                  Yoshua Bengio},
  editor       = {Yoshua Bengio and
                  Yann LeCun},
  title        = {Neural Machine Translation by Jointly Learning to Align and Translate},
  booktitle    = {3rd International Conference on Learning Representations, {ICLR} 2015,
                  San Diego, CA, USA, May 7-9, 2015, Conference Track Proceedings},
  year         = {2015},
  url          = {http://arxiv.org/abs/1409.0473},
  bibsource    = {dblp computer science bibliography, https://dblp.org}
}

@article{Google2016MultilingualNMT,
  author    = {Melvin Johnson and
               Mike Schuster and
               Quoc V. Le and
               Maxim Krikun and
               Yonghui Wu and
               Zhifeng Chen and
               Nikhil Thorat and
               Fernanda B. Vi{\'{e}}gas and
               Martin Wattenberg and
               Greg Corrado and
               Macduff Hughes and
               Jeffrey Dean},
  title     = {Google's Multilingual Neural Machine Translation System: Enabling
               Zero-Shot Translation},
  journal   = {CoRR},
  volume    = {abs/1611.04558},
  year      = {2016},
  url       = {http://arxiv.org/abs/1611.04558},
  bibsource = {dblp computer science bibliography, http://dblp.org}
}

@inproceedings{han-etal-2020-multimwe,
    title = "{M}ulti{MWE}: Building a Multi-lingual Multi-Word Expression ({MWE}) Parallel Corpora",
    author = "Han, Lifeng  and
      Jones, Gareth  and
      Smeaton, Alan F.",
    booktitle = "Proceedings of the 12th Language Resources and Evaluation Conference",
    month = may,
    year = "2020",
    address = "Marseille, France",
    publisher = "European Language Resources Association",
    url = "https://www.aclweb.org/anthology/2020.lrec-1.363",
    pages = "2970--2979",
    language = "English",
    ISBN = "979-10-95546-34-4",
}

@InProceedings{Radical15ShiNLP,
  author = 	"Shi, Xinlei
		and Zhai, Junjie
		and Yang, Xudong
		and Xie, Zehua
		and Liu, Chao",
  title = 	"Radical Embedding: Delving Deeper to Chinese Radicals",
  booktitle = 	"Proceedings of the 53rd Annual Meeting of the Association for      Computational Linguistics and the 7th International Joint Conference on      Natural Language Processing (Volume 2: Short Papers)    ",
  year = 	"2015",
  publisher = 	"Association for Computational Linguistics",
  pages = 	"594--598",
  location = 	"Beijing, China",
  doi = 	"10.3115/v1/P15-2098",
  url = 	"http://aclanthology.coli.uni-saarland.de/pdf/P/P15/P15-2098.pdf"
}

@inproceedings{Pinnis2013mp_aligner,
    title = "Context Independent Term Mapper for {E}uropean Languages",
    author = "Pinnis, M{\=a}rcis",
    booktitle = "Proceedings of the International Conference Recent Advances in Natural Language Processing {RANLP} 2013",
    month = sep,
    year = "2013",
    address = "Hissar, Bulgaria",
    publisher = "INCOMA Ltd. Shoumen, BULGARIA",
    url = "https://www.aclweb.org/anthology/R13-1074",
    pages = "562--570",
}

@inproceedings{SubwordNMT15Sennrich,
    title = "Neural Machine Translation of Rare Words with Subword Units",
    author = "Sennrich, Rico  and
      Haddow, Barry  and
      Birch, Alexandra",
    booktitle = "Proceedings of the 54th Annual Meeting of the Association for Computational Linguistics (Volume 1: Long Papers)",
    month = aug,
    year = "2016",
    address = "Berlin, Germany",
    publisher = "Association for Computational Linguistics",
    url = "https://www.aclweb.org/anthology/P16-1162",
    doi = "10.18653/v1/P16-1162",
    pages = "1715--1725",
}

@inproceedings{erofeev-etal-2021-cushlepor,
    title = "cush{LEPOR} uses {LABSE} distilled knowledge to improve correlation with human translations",
    author = "Erofeev, Gleb  and
      Sorokina, Irina  and
      Han, Lifeng  and
      Gladkoff, Serge",
    booktitle = "Proceedings of Machine Translation Summit XVIII: Users and Providers Track",
    month = aug,
    year = "2021",
    address = "Virtual",
    publisher = "Association for Machine Translation in the Americas",
    url = "https://aclanthology.org/2021.mtsummit-up.28",
    pages = "421--439",
}

@inproceedings{koehn-knowles-2017-six,
    title = "Six Challenges for Neural Machine Translation",
    author = "Koehn, Philipp  and
      Knowles, Rebecca",
    booktitle = "Proceedings of the First Workshop on Neural Machine Translation",
    month = aug,
    year = "2017",
    address = "Vancouver",
    publisher = "Association for Computational Linguistics",
    url = "https://aclanthology.org/W17-3204",
    doi = "10.18653/v1/W17-3204",
    pages = "28--39",
}

@book{han2014lepor,
  title={LEPOR: An Augmented Machine Translation Evaluation Metric},
  author={Han, Lifeng},
  year={2014},
  url= {http://arxiv.org/abs/1703.08748},
  publisher={University of Macau, MSc. Thesis},
}

@inproceedings{han2012lepor,
  title={LEPOR: A Robust Evaluation Metric for Machine Translation with Augmented Factors},
  author={Han, Lifeng and Wong, Derek F. and Chao, Lidia S.},
  booktitle={Proceedings of the 24th International Conference on Computational Linguistics (COLING 2012)},
  pages={Posters--441},
  year={2012},
  organization={Association for Computational Linguistics}
}

@InProceedings{Sag2002MWE,
author="Sag, Ivan A.
and Baldwin, Timothy
and Bond, Francis
and Copestake, Ann
and Flickinger, Dan",
editor="Gelbukh, Alexander",
title="Multiword Expressions: A Pain in the Neck for NLP",
booktitle="Computational Linguistics and Intelligent Text Processing",
year="2002",
publisher="Springer Berlin Heidelberg",
address="Berlin, Heidelberg",
pages="1--15",
isbn="978-3-540-45715-2"
}

@inproceedings{Bouamor2012IdentifyingBM,
  title={Identifying bilingual Multi-Word Expressions for Statistical Machine Translation},
  author={Dhouha Bouamor and Nasredine Semmar and Pierre Zweigenbaum},
  booktitle={Conference on Language Resources and Evaluation},
  year={2012}
}

@inproceedings{lambert2005mwe,
  TITLE = {{Data Inferred Multi-word Expressions for Statistical Machine Translation}},
  AUTHOR = {Lambert, Patrik and Banchs, Rafael E.},
  URL = {https://hal.archives-ouvertes.fr/hal-00705007},
  BOOKTITLE = {{Proceedings of Machine Translation Summit X}},
  ADDRESS = {Phuket, Thailand},
  PAGES = {396-403},
  YEAR = {2005},
  MONTH = Sep,
  HAL_ID = {hal-00705007},
  HAL_VERSION = {v1},
}

@inproceedings{wmt2017findings,
    title = "Findings of the 2017 Conference on Machine Translation ({WMT}17)",
    author = "Bojar, Ond{\v{r}}ej  and
      Chatterjee, Rajen  and
      Federmann, Christian  and
      Graham, Yvette  and
      Haddow, Barry  and
      Huang, Shujian  and
      Huck, Matthias  and
      Koehn, Philipp  and
      Liu, Qun  and
      Logacheva, Varvara  and
      Monz, Christof  and
      Negri, Matteo  and
      Post, Matt  and
      Rubino, Raphael  and
      Specia, Lucia  and
      Turchi, Marco",
    booktitle = "Proceedings of the Second Conference on Machine Translation",
    month = sep,
    year = "2017",
    address = "Copenhagen, Denmark",
    publisher = "Association for Computational Linguistics",
    url = "https://www.aclweb.org/anthology/W17-4717",
    doi = "10.18653/v1/W17-4717",
    pages = "169--214",
}

@InProceedings{wmt2018findings,
  author    = {Bojar, Ond{\v{r}}ej  and  Federmann, Christian  and  Fishel, Mark  and  Graham, Yvette  and  Haddow, Barry  and  Huck, Matthias  and  Koehn, Philipp  and  Monz, Christof},
  title     = {Findings of the 2018 Conference on Machine Translation {(WMT18)}},
  booktitle = {Proceedings of the Third Conference on Machine Translation, Volume 2: Shared Task Papers},
  month     = {October},
  year      = {2018},
  address   = {Belgium, Brussels},
  publisher = {Association for Computational Linguistics},
  pages     = {272--307},
  url       = {http://www.aclweb.org/anthology/W18-6401}
}

@inproceedings{Ren2009mwe,
 author = {Ren, Zhixiang and L\"{u}, Yajuan and Cao, Jie and Liu, Qun and Huang, Yun},
 title = {Improving Statistical Machine Translation Using Domain Bilingual Multiword Expressions},
 booktitle = {Proceedings of the Workshop on Multiword Expressions: Identification, Interpretation, Disambiguation and Applications},
 series = {MWE '09},
 year = {2009},
 isbn = {978-1-932432-60-2},
 location = {Suntec, Singapore},
 pages = {47--54},
 numpages = {8},
 url = {http://dl.acm.org/citation.cfm?id=1698239.1698249},
 acmid = {1698249},
 publisher = {Association for Computational Linguistics},
 address = {Stroudsburg, PA, USA},
}

@inproceedings{schmid1994treetagger,
  address = {Manchester, UK},
  author = {Schmid, Helmut},
  booktitle = {Proceedings of the International Conference on New Methods in Language
	Processing},
  title = {Probabilistic Part-of-Speech Tagging Using Decision Trees},
  year = 1994
}

@book{ramisch2015book_mwetoolkit,
	category = {OS},
	doi = {10.1007/978-3-319-09207-2},
	volume = {XIV},
	year = {2015},
	pages = {230},
	isbn = {978-3-319-09206-5},
	series = {Theory and Applications of Natural Language Processing},
	publisher = {Springer},
	title = {Multiword Expressions Acquisition: {A} Generic and Open Framework},
	author = {Carlos Ramisch},
}

\section{Language Resource References}
\label{lr:ref}

\begin{itemize}
    \item \textsc{AlphaMWE} multilingual parallel corpora: annotated by human annotators with MWEs, including Chinese, English, Polish, German, Arabic languages \cite{hantowards}. English, Chinese, German, and Polish have 750 sentences \cite{han-etal-2020-alphamwe}; Arabic version has 150 segments including morden standard Arabic (MSA), Egyptian and Moroccan dialectal Arabic \cite{hadj-mohamed-etal-2023-alphamwe}.
    \item \textsc{MultiMWE} parallel corpora: automatically extracted from bilingual corpora using MT assistance and filtering, including English-Chinese and English-German pairs \cite{han-etal-2020-multimwe}.
\end{itemize}

\bibliographystylelanguageresource{lrec2026-natbib}
\bibliographylanguageresource{languageresource}


\end{document}